\documentclass{article}

\PassOptionsToPackage{numbers, compress}{natbib}

\usepackage[final]{neurips_2024}

\usepackage[utf8]{inputenc} 
\usepackage[T1]{fontenc}    
\usepackage[colorlinks]{hyperref}
\usepackage{url}            
\usepackage{booktabs}       
\usepackage{amsfonts}       
\usepackage{nicefrac}       
\usepackage{microtype}      
\usepackage{xcolor}         

\usepackage{multirow}
\usepackage{graphicx}
\usepackage{caption,setspace}
\usepackage{subfig}
\usepackage{amsmath}
\usepackage{bm}
\usepackage{wrapfig}
\usepackage{bbding}

\definecolor{blue}{rgb}{0.22, 0.22, 0.95}
\definecolor{lightgray}{RGB}{180,180,180}
\definecolor{baselinecolor}{gray}{.9}

\makeatletter
\usepackage{xspace}
\def\@onedot{\ifx\@let@token.\else.\null\fi\xspace}
\DeclareRobustCommand\onedot{\futurelet\@let@token\@onedot}

\newcommand{\tabref}[1]{Tab\onedot~\ref{#1}}

\def\eg{\emph{e.g}\onedot} 
\def\ie{\emph{i.e}\onedot} 
 
 \def\vs{\emph{vs}\onedot}

\newcommand{\figref}[1]{Fig\onedot~\ref{#1}}
\newcommand{\secref}[1]{Sec\onedot~\ref{#1}}

\newlength\savewidth\newcommand\shline{\noalign{\global\savewidth\arrayrulewidth
  \global\arrayrulewidth 1pt}\hline\noalign{\global\arrayrulewidth\savewidth}}
\newcommand{\tablestyle}[2]{\setlength{\tabcolsep}{#1}\renewcommand{\arraystretch}{#2}\centering\footnotesize}

\newcommand{\modelname}{DiMR\xspace}
\newcommand{\tdpname}{TD-LN\xspace}

\title{Alleviating Distortion in Image Generation via Multi-Resolution Diffusion Models and Time-Dependent Layer Normalization}

\author{
  \hspace{-1ex}Qihao Liu\textsuperscript{1,2*}, Zhanpeng Zeng\textsuperscript{1,3*}, Ju He\textsuperscript{1,2*}, Qihang Yu\textsuperscript{1}, Xiaohui Shen\textsuperscript{1}, Liang-Chieh Chen\textsuperscript{1}
  \vspace{2px}
  \\
  \vspace{2px}
  \textsuperscript{1} ByteDance \qquad
   \textsuperscript{2} Johns Hopkins University \qquad
   \textsuperscript{3} University of Wisconsin-Madison \qquad \\
   * equal contribution \\
   \url{https://qihao067.github.io/projects/DiMR}
}

\begin{document}

\maketitle

{%
\begin{center}
    \centering
    \captionsetup{type=figure}
    \includegraphics[width=\linewidth]{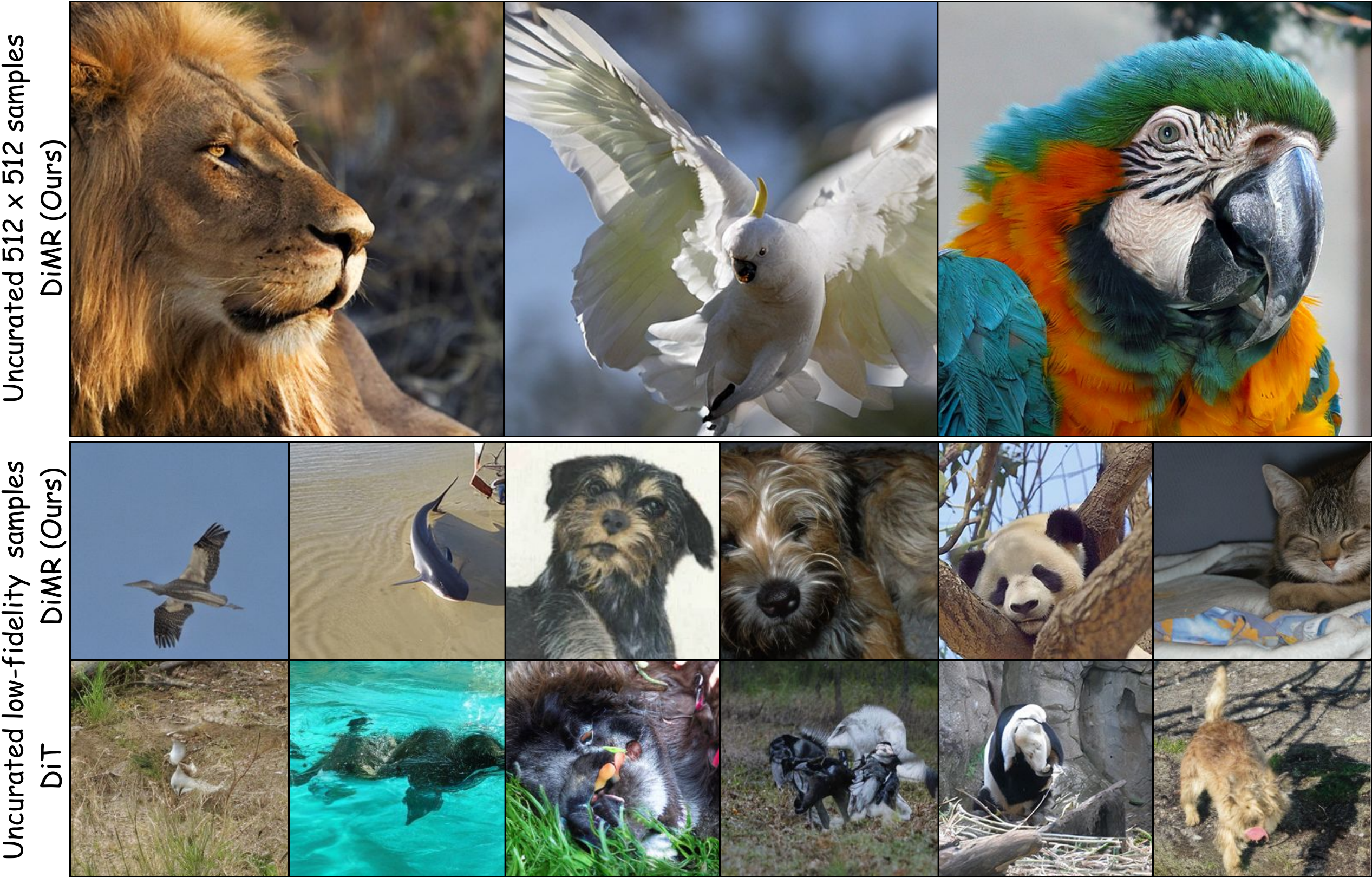}
    \captionof{figure}{
    (Top) Randomly sampled $512\times512$ images generated by the proposed \modelname. (Bottom) Random samples of the \textbf{low visual fidelity} $256\times256$ images generated by  \modelname and DiT~\cite{peebles2023scalable}. 
    To detect low visual fidelity images for both models, a classifier-based rejection model is employed (with the same rejection rate). 
    \modelname generates images with higher fidelity and less distortion than DiT.
    }
    \label{fig:teaser}
\end{center}%
}

\begin{abstract}
This paper presents innovative enhancements to diffusion models by integrating a novel multi-resolution network and time-dependent layer normalization.
Diffusion models have gained prominence for their effectiveness in high-fidelity image generation.
While conventional approaches rely on convolutional U-Net architectures, recent Transformer-based designs have demonstrated superior performance and scalability.
However, Transformer architectures, which tokenize input data (via ``patchification''), face a trade-off between visual fidelity and computational complexity due to the quadratic nature of self-attention operations concerning token length.
While larger patch sizes enable attention computation efficiency, they struggle to capture fine-grained visual details, leading to image distortions.
To address this challenge, we propose augmenting the \textbf{Di}ffusion model with the \textbf{M}ulti-\textbf{R}esolution network (\modelname), a framework that refines features across multiple resolutions, progressively enhancing detail from low to high resolution.
Additionally, we introduce Time-Dependent Layer Normalization (\tdpname), a parameter-efficient approach that incorporates time-dependent parameters into layer normalization to inject time information and achieve superior performance.
Our method's efficacy is demonstrated on the class-conditional ImageNet generation benchmark, where \modelname-XL variants surpass previous diffusion models, achieving FID scores of 1.70 on ImageNet $256 \times 256$ and 2.89 on ImageNet $512 \times 512$. Our best variant, \modelname-G, further establishes a state-of-the-art 1.63 FID on ImageNet $256 \times 256$.
\end{abstract}

\section{Introduction}
\label{sec:intro}

Diffusion and score-based generative models~\cite{hyvarinen2005estimation,sohl2015deep,song2019generative,ho2020denoising,song2020denoising} have demonstrated promising results for high-fidelity image generation~\cite{dhariwal2021diffusion,nichol2021glide,ramesh2022hierarchical,rombach2022high,saharia2022photorealistic}.
These models generate images through an iterative process of gradually denoising Gaussian random noise to create realistic samples
Central to this process is a neural network, tasked with denoising the inputs through a mean squared error loss function. 
Traditionally, U-Net architectures~\cite{ronneberger2015u} (enhanced with residual blocks~\cite{he2016deep} and self-attention blocks~\cite{vaswani2017attention} at lower resolution) have been prevalent. However, recent advancements have introduced Transformer-based designs~\cite{vaswani2017attention,dosovitskiy2020image}, offering superior performance and scalability.

In practice, Transformer-based architectures face the challenge of balancing visual fidelity and computational complexity, primarily stemming from the self-attention operation and the patchification process employed for downsampling inputs~\cite{dosovitskiy2020image} (\ie, a smaller patch size results in better visual fidelity at the cost of a longer token length and thus more computational complexity by the self-attention operation).
The quadratic complexity inherent in self-attention concerning token length necessitates larger patch sizes to facilitate more efficient attention computations.
However, the adoption of large patch sizes inevitably compromises the model's capacity to capture finer visual details, resulting in image distortion (\ie, low visual fidelity). 
This dilemma prompts DiT~\cite{peebles2023scalable} to conduct a systematic study on the impact of patch size on image distortion, as depicted in Fig.~7 of their paper.
Consequently, they settled on a patch size of 2 for their final design.
Similarly, U-ViT~\cite{bao2023all} opted for a patch size of 2 for input sizes of $256\times256$ and a patch size of 4 for $512\times512$ images, effectively balancing the token length for different image sizes.
Despite these meticulous adjustments, the generated results still exhibit discernible image distortion, as illustrated in Fig.~\ref{fig:teaser}.


One simplistic solution to mitigate image distortion in Transformer-based architectures is adopting a patch size of 1, but this significantly increases computational complexity.
Instead, inspired by the success of \textit{image cascade}~\cite{ho2022cascaded, saharia2022photorealistic} which generate images at increasing resolutions, we propose a \textit{feature cascade} approach that progressively upsamples lower-resolution features to higher resolutions, alleviating distortion in image generation.
In this study, we present \modelname, which enhances the \textbf{Di}ffusion model with a \textbf{M}ulti-\textbf{R}esolution network. 
\modelname tackles the challenge of balancing visual detail capture and computational complexity through improvements in the denoising backbone architecture.
We employ a multi-resolution network design that comprises multiple branches to progressively refine features from low to high resolution, preserving intricate details within the input data.
Specifically, the first branch handling the lowest resolution incorporates Transformer blocks~\cite{vaswani2017attention}, leveraging the superior performance and scalability observed in prior works~\cite{bao2023all,peebles2023scalable}, while the remaining branches utilize ConvNeXt blocks~\cite{liu2022convnet}, which are efficient for high resolution features.
The network processes inputs progressively from the lowest resolution, with additional features from the preceding resolution.
The last branch refines features at the same spatial resolution as the input, effectively mitigating image distortion arising from the patchification.

Additionally, we observe that existing time conditioning mechanisms~\cite{perez2018film, karras2019style, dhariwal2021diffusion}, such as adaptive layer normalization (adaLN)~\cite{peebles2023scalable}, are parameter-intensive. In contrast, we propose a more efficient approach, Time-Dependent Layer Normalization (\tdpname), that integrates time-dependent parameters directly into layer normalization~\cite{ba2016layer}, achieving superior performance with fewer parameters.

To demonstrate its effectiveness, we evaluate \modelname on the class-conditional ImageNet generation benchmark~\cite{deng2009imagenet}.
On ImageNet $64\times64$, \modelname-M (133M parameters) and \modelname-L (284M), without classifier-free guidance~\cite{ho2022classifier}, achieve FID scores of 3.65 and 2.21, respectively, outperforming the Transformer-based U-ViT-M/4 and U-ViT-L/4 by 2.20 and 2.05 FID.
On ImageNet $256\times256$, \modelname-XL (505M) achieves FID scores of 4.50 without classifier-free guidance and 1.70 with classifier-free guidance. Meanwhile, \modelname-G (1.06B) further improves the FID scores to 3.56 without classifier-free guidance and 1.63 with classifier-free guidance.
On ImageNet $512\times512$, \modelname-XL (525M) achieves FID scores of 7.93 and 2.89, without and with classifier-free guidance, respectively.
These results demonstrate superior performance compared to all previous methods, despite having similar or smaller model sizes, establishing a new state-of-the-art performance. In summary, our main contributions are as follows:
\begin{enumerate}
    \item We develop effective strategies for integrating multi-resolution networks into diffusion models, introducing the novel feature cascade approach that captures visual details and reduces image distortions in high-fidelity image generation.
    \item We propose \tdpname, a simple yet effective parameter-efficient method that explicitly encodes crucial temporal information into the diffusion model for enhanced performance.
    \item We introduce \modelname, a novel architecture that enhances diffusion models with the proposed multi-resolution network and the \tdpname. \modelname demonstrates superior performance on the class-conditional ImageNet generation benchmark compared to existing methods.
\end{enumerate}
\section{Related Work}
\label{sec:related_work}

\textbf{Diffusion models.}
Diffusion~\citep{sohl2015deep,ho2020denoising} and score-based generative models~\citep{hyvarinen2005estimation,song2019generative}, centered around a denoising network trained to progressively produce denoised variants of the input data.
They have driven significant advances across various domains~\citep{li2022diffusion,kong2020diffwave,vahdat2022lion,tashiro2021csdi,xu2022geodiff,nie2022diffusion,xu2023open}, particularly excelling in high-fidelity image generation tasks~\citep{nichol2021glide,ramesh2022hierarchical,rombach2022high,saharia2022photorealistic}.
Key advancements in diffusion models include the improvements in sampling methodologies~\citep{ho2020denoising,song2020denoising,karras2022elucidating} and the adoption of classifier-free guidance~\citep{ho2022classifier}.
Latent Diffusion Models (LDMs)~\citep{rombach2022high,peebles2023scalable,podell2023sdxl,xue2023raphael} address the challenges of high-resolution image generation by conducting diffusion in the lower-resolution latent space via a pre-trained autoencoder~\cite{kingma2013auto}.
In this study, our focus lies on designing the denoising network within diffusion models and examining its applicability across both pixel diffusion models and LDMs.

\textbf{Architecture for diffusion models.}
Early diffusion models employed convolutional U-Net architectures~\cite{ronneberger2015u} as the denoising network, which were subsequently strengthened through explorations of either computing attention~\cite{dhariwal2021diffusion,nichol2021improved} or performing diffusion directly at multiple scales~\cite{ho2022cascaded,gu2023matryoshka}.
Recently, Transformer-based architectures~\cite{bao2023all,peebles2023scalable,hatamizadeh2023diffit} along with other explorations~\cite{yan2023diffusion,tian2024visual,kim2024pagoda} have emerged as promising alternatives, showcasing superior performance and scalability. Specifically, for Transformer-based architectures, U-ViT~\cite{bao2023all} treats all inputs, including time, condition, and noisy image patches, as tokens and employs long-skip connections between shallow and deep transformer layers inspired by U-Net. Similarly, DiT~\cite{peebles2023scalable} leverages Vision Transformers (ViTs)~\cite{dosovitskiy2020image} to systematically explore the design space under the Latent Diffusion Models (LDMs) framework, demonstrating favorable properties such as scalability, robustness, and efficiency.
In this study, we introduce the Multi-Resolution Network as a new denoising architecture for diffusion models, featuring a multi-branch design where each branch is dedicated to processing a specific resolution.


\textbf{Time conditioning mechanisms.}
Following the widespread usage of adaptive normalization~\cite{perez2018film} in GANs~\cite{brock2018large,karras2019style}, diffusion models similarly explore adaptive group normalization (AdaGN)~\cite{dhariwal2021diffusion} and adaptive layer normalization (AdaLN)~\cite{peebles2023scalable} to encode the time information. These methods share the similarity in requiring computing a linear projection of the timestep, which significantly increases the parameter of the model. Recently, U-ViT~\cite{bao2023all} introduces a new strategy to simply treat time as a token and process with Transformer blocks. Even though effective, it is not feasible to treat time as input for other blocks (\eg, ConvNeXt blocks~\cite{liu2022convnet}).
In this study, we introduce Time-Dependent Layer Normalization (\tdpname), a parameter-efficient approach that explicitly encodes temporal information by incorporating time-dependent parameters into layer normalization~\cite{ba2016layer}.

\section{Preliminary}
\label{sec:preliminary}
\textbf{Diffusion models~\citep{sohl2015deep,ho2020denoising}} are characterized by a forward process that gradually injects noises to destroy data $\bm{x}_0 \sim q(\bm{x}_0)$, and a reverse process that inverts the forward process corruptions.
Formally, the noise injection process is formulated as a Markov chain:
\begin{align*}
    q(\bm{x}_{1:T}|\bm{x}_0) = \prod_{t=1}^T q(\bm{x}_t|\bm{x}_{t-1}),
\end{align*}
where $\bm{x}_t$ for $t \in [1:T]$ is a family of random variables obtained by progressively injecting Gaussian noise into the data  $\bm{x}_0$, and $q(\bm{x}_t|\bm{x}_{t-1}) = \mathcal{N}(\bm{x}_t|\sqrt{\alpha_t} \bm{x}_{t-1}, \beta_t \bm{I})$ represents the noise injection schedule such that $\alpha_t + \beta_t = 1$.
In the reverse process, a Gaussian model $p(\bm{x}_{t-1}|\bm{x}_t) = \mathcal{N}(\bm{x}_{t-1}|\bm{\mu}_t(\bm{x}_t), \sigma_t^2 \bm{I})$ is learned to approximate the ground truth reverse transition $q(\bm{x}_{t-1}|\bm{x}_t)$. This step is equivalent to predicting the denoised variant of the input $\bm{x}_t$, and thus the learning objective can be further simplified to predicting the noise $\bm{\epsilon}_t$ via a noise prediction network (with parameters $\bm{\theta}$), \ie, $\bm{\epsilon}_{\bm{\theta}}(\bm{x}_t)$: $\min\limits_{\bm{\theta}} \mathbb{E}_{t, \bm{x}_0, \bm{\epsilon}_t} \| \bm{\epsilon}_t - \bm{\epsilon}_{\bm{\theta}}(\bm{x}_t) \|_2^2$. The condition information $c$ can be incorporated into the learning objective when the diffusion process is guided by the class condition~\citep{dhariwal2021diffusion}, \ie,
$\bm{\epsilon}_{\bm{\theta}}(\bm{x}_t, c)$: $\min\limits_{\bm{\theta}} \mathbb{E}_{t, \bm{x}_0, c, \bm{\epsilon}_t} \| \bm{\epsilon}_t - \bm{\epsilon}_{\bm{\theta}}(\bm{x}_t, c) \|_2^2$. 
Traditionally, learning this objective relies on the U-Net~\cite{ronneberger2015u}, with the condition $c$ encoded into the U-Net through various methods~\cite{bao2023all,dhariwal2021diffusion,peebles2023scalable,rombach2022high}.


\textbf{Classifier-free guidance~\cite{ho2022classifier}}, an effective approach to generating high-fidelity samples, combines the score estimates from a conditional diffusion model and a jointly trained unconditional diffusion model.
Formally, the classifier-free guidance encourages the sampled $\bm{x}$ to have 
high $p(\bm{x}|c)$ by setting:
$\hat{\epsilon}_\theta(\bm{x}_t, c) = \epsilon_\theta(\bm{x}_t,\emptyset) + s \cdot \nabla_{\bm{x}} \log p(\bm{x}|c) \propto \epsilon_\theta(\bm{x}_t, \emptyset) + s \cdot (\epsilon_\theta(\bm{x}_t, c) - \epsilon_\theta(\bm{x}_t, \emptyset))$, where $s$ is the scale of guidance, $s\geq1$, and setting $s=1$ becomes the standard sampling. 
Following prior arts~\cite{bao2023all,peebles2023scalable}, we also exploit classifier-free guidance.



\section{Method}
\label{sec:method}
In this section, we begin by introducing the proposed Multi-Resolution Network (\secref{sec:multi_res}), which progressively refines features from low to high resolution.
Next, we detail the proposed Time-Dependent Layer Normalization (\secref{sec:tdln}).
We then discuss several micro-level design enhancements (\secref{sec:micro}).
Finally, we present the \modelname model variants, scaled for different model sizes (\secref{sec:meta_arch}).

\subsection{Multi-Resolution Network}
\label{sec:multi_res}

\begin{figure*}
    \centering
    \includegraphics[width=0.95\linewidth]{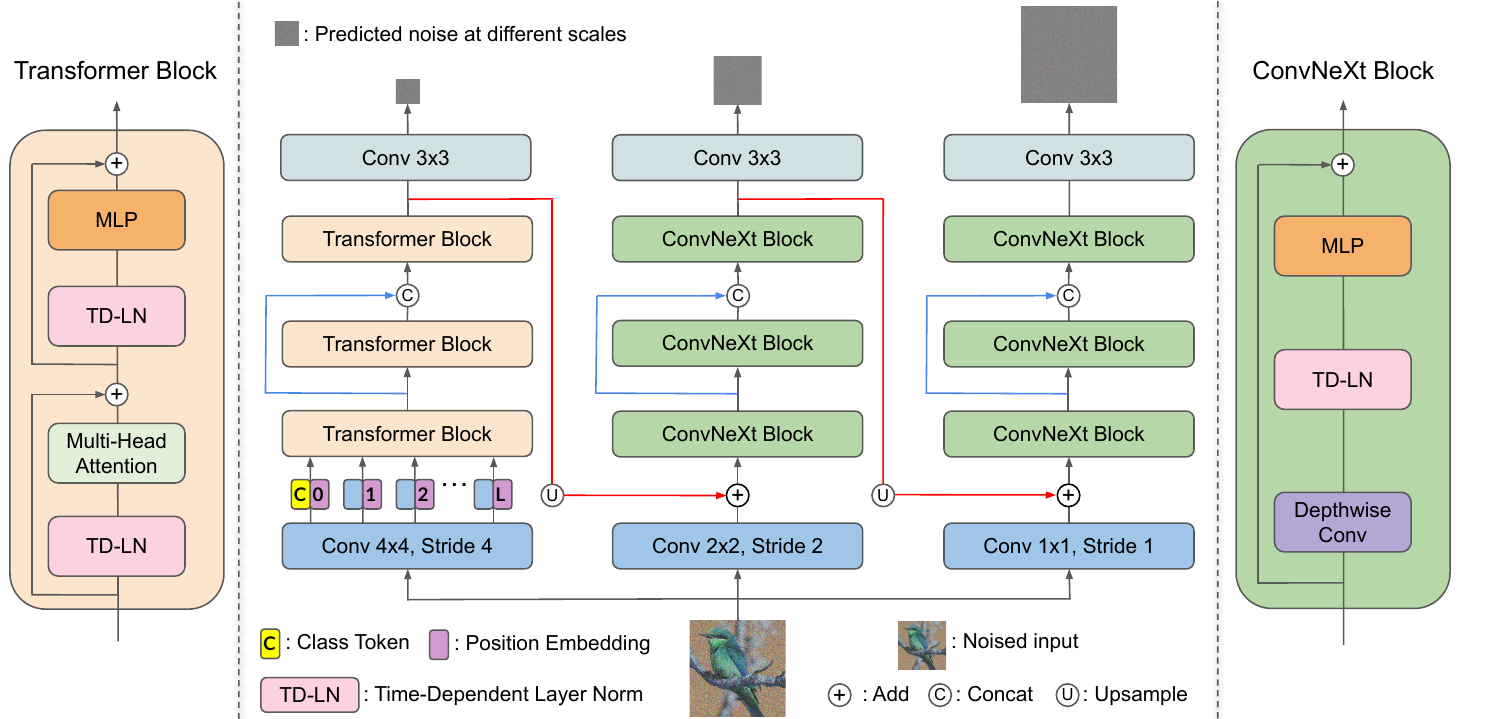}
    \caption{\textbf{Model overview.}
    We propose \modelname that enhances \textbf{D}iffusion models with a \textbf{M}ulti-\textbf{R}esolution Network.
    In the figure, we present the Multi-Resolution Network with three branches.
    The first branch processes the lowest resolution (4 times smaller than the input size) using powerful Transformer blocks, while the other two branches handle higher resolutions (2 times smaller than the input size and the same size as the input, respectively) using effective ConvNeXt blocks. The network employs a \textit{feature cascade} framework, progressively upsampling lower-resolution features to higher resolutions to reduce distortion in image generation. The Transformer and ConvNeXt blocks are further enhanced by the proposed Time-Dependent Layer Normalization (TD-LN), detailed in~\figref{fig:td-ln}.
    }
    \label{fig:model_arch}
\end{figure*}

\textbf{Motivation.}
There is a trade-off between generation quality and computational complexity as depicted in the ablation study in Fig.~7 of DiT~\cite{peebles2023scalable}. 
Their careful study revealed that Transformer-based diffusion models with smaller patch sizes operate at higher feature resolutions and produce better generation quality but incur higher computational costs due to the increased input size. 

We conjecture that the distortion in U-ViT~\cite{bao2023all} and DiT~\cite{peebles2023scalable} arises from their oversimplified upsampling module, where lower-resolution feature maps are upsampled directly to the target size of the generated images via a simple linear layer (for increasing channels) and pixel shuffling upsampling~\cite{shi2016real}.
Inspired by \textit{image cascade}~\cite{ho2022cascaded, saharia2022photorealistic}—a method for generating high-resolution images by using multiple cascaded diffusion models to produce images of progressively increasing resolution—we propose \textit{feature cascade}, which progressively upsample lower-resolution features to higher resolutions to alleviate distortion in image generation. The \textit{feature cascade} is implemented through the proposed Multi-Resolution Network, deployed as the denoising network in diffusion models.

\textbf{Overview of multi-branch design.}
The proposed Multi-Resolution Network comprises $R$ branches, where each branch is dedicated to process a specific resolution.
For the $r$-th branch ($r \in \{1, \cdots, R\}$), the input features are processed by a convolution with a kernel size of $2^{R-r}\times2^{R-r}$ and a stride of $2^{R-r}$, which effectively patchifies the input for different resolutions.
The first branch (\ie, $r=1$) downsamples the input features by a factor of $2^{R-1}$, and subsequently handles the lowest resolution features via the Transformer blocks~\cite{vaswani2017attention}, which enjoys the superior performance and scalability of self-attention operations~\cite{bao2023all,peebles2023scalable}. 
For higher resolution features, the remaining branches utilize the ConvNeXt blocks~\cite{liu2022convnet}, which leverages the efficiency of large kernel depthwise-convolution operations~\cite{howard2017mobilenets,sandler2018mobilenetv2}.
Intermediate features from the previous branch (last block's output) are upsampled and added with the inputs for the current branch.
Following U-ViT~\cite{bao2023all}, all branches employ the long skip connections and an additional $3\times3$ convolution in the end.
The final branch (\ie, $r=R$) refines features at the same spatial resolution as the input. 

\textbf{Design details.}
For $r \in \{1, \cdots, R\}$, we define the $r$-th branch as a function $f_{\bm{\theta}, r}$ as follows:
\begin{equation}
\bm{\epsilon}_{\bm{\theta}}(\bm{x}_t, c, r), \bm{y}_r = f_{\bm{\theta}, r}(\bm{x}_t, \bm{y}_{r-1}, t, c),
\end{equation}
where the function $f_{\bm{\theta}, r}$, parameterized by $\bm{\theta}$ and $r$, takes as input the input features $\bm{x}_t$ and the features from previous resolution $\bm{y}_{r-1}$ (also time $t$ and condition $c$). The outputs of $f_{\bm{\theta}, r}$ contain the intermediate features $\bm{y}_r$ (last block's output, before the final $3\times3$ convolution) and the predicted noise $\bm{\epsilon}_{\bm{\theta}}(\bm{x}_t, c, r)$ for the resolution specific to $r$-th branch.

To process the inputs, the function $f_{\bm{\theta}, r}$ first patchifies the input features $\bm{x}_t$, and adds it with the upsampled features $\bm{y}_{r-1}$ from the previous resolution $r-1$. The resulting features are then processed by either a stack of Transformer blocks (when $r=1$) or ConvNeXt blocks (when $r\neq1$) with another $3\times3$ convolution added in the end. Formally, we have:
\begin{equation}
    f_{\bm{\theta}, r}(\bm{x}_t, \bm{y}_{r-1}, t, c) = \text{Conv}_{3\times3}(g_{\bm{\theta}, r}(\text{Patchify}(\bm{x}_t) + \text{Upsample}(\bm{y}_{r-1}), t, c)),
\end{equation}
where $\text{Conv}_{3\times3}$ is $3\times3$ convolution, $\text{Patchify}$ is patchification instantiated via a convolution with a kernel size of $2^{R-r} \times 2^{R-r}$ and a stride of $2^{R-r}$, $\text{Upsample}$ is the pixel shuffling upsampling operation~\cite{shi2016real}, and $g_{\bm{\theta}, r}$ is a stack of Transformer blocks or ConvNeXt blocks, depending on $r$, augmented with the long skip connections~\cite{bao2023all}.
For the first branch (\ie, $r=1$), $\bm{y}_0$ is set to zero. The noise prediction  $\bm{\epsilon}_{\bm{\theta}}(\bm{x}_t, c, R)$ at the last branch (\ie, $r=R$) is used for the iterative diffusion process. We illustrate the proposed Multi-Resolution Network with three branches in~\figref{fig:model_arch}.
Note that the input features can be either raw image pixels or latent features after VAE~\cite{kingma2013auto}, where the latent features facilitate efficient high-resolution image generation~\cite{rombach2022high}.

\subsection{Time-Dependent Layer Normalization}
\label{sec:tdln}

\begin{figure*}
    \centering
    \includegraphics[width=0.95\linewidth]{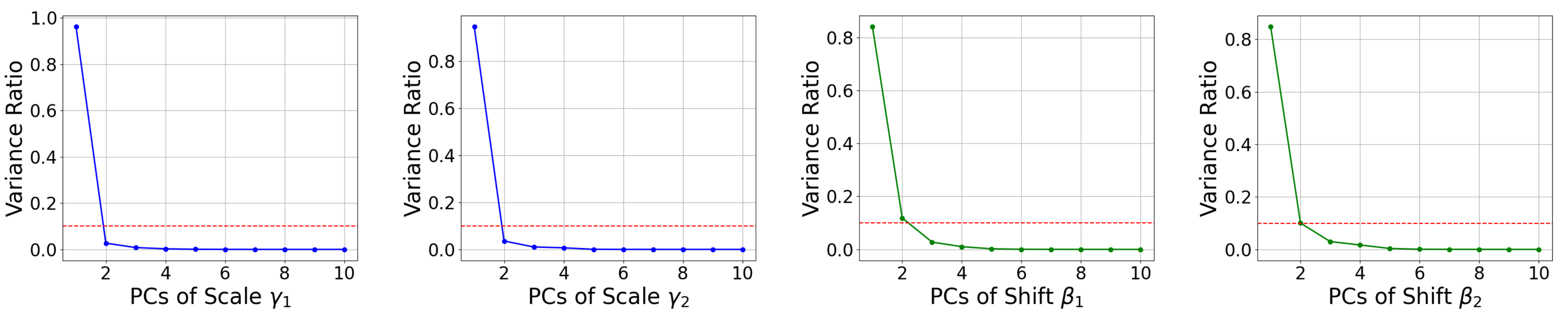}
    \caption{\textbf{Principal Component Analysis (PCA) of learned scale and shift parameters in adaLN-Zero~\cite{peebles2023scalable}.} We conduct PCA on the learned scale ($\gamma_1$, $\gamma_2$) and shift ($\beta_1$, $\beta_2$) parameters obtained from a parameter-heavy MLP in adaLN-Zero using a pre-trained DiT-XL/2~\cite{peebles2023scalable} model. The vertical axis represents the explained variance ratio of the corresponding Principal Components (PCs). Our observations reveal that the learned parameters can be largely explained by two principal components, suggesting the potential to approximate them by a simpler function.}
    \label{fig:adaln}
\end{figure*}

\textbf{Motivation.}
Time conditioning plays a crucial role in the diffusion process. While the ConvNeXt blocks in the Multi-Resolution Network efficiently process high-resolution features, they also present a new challenge: \textit{How do we inject time information into ConvNeXt blocks?}
To address this, we carry out a systematic ablation study (details in Tab.~\ref{tab:ablation}), starting with the U-ViT architecture~\cite{bao2023all}, which encodes time information via an in-context conditioning mechanism, Time-Token (\ie, treating time as input token to Transformer).
Unlike Transformer blocks, however, it is not feasible to add a time token directly to ConvNeXt blocks, which can only process 2D features.
As an alternative, we explored the adaptive normalization mechanism, particularly the adaptive layer normalization AdaLN-Zero~\cite{peebles2023scalable}.
Interestingly, we found AdaLN-Zero to be more effective than Time-Token on the ImageNet $64\times64$ benchmark, contradicting U-ViT's findings on CIFAR-10~\cite{krizhevsky2009learning} (See Fig.~2(b) in~\cite{bao2023all}).
However, AdaLN-Zero significantly increases model parameters (from 130.9M to 202.4M) due to the Multi-Layer Perceptron (MLP) used to adaptively learn the scale and shift parameters.
\begin{wrapfigure}{l}{0.42\textwidth}
    \includegraphics[width=0.42\textwidth]{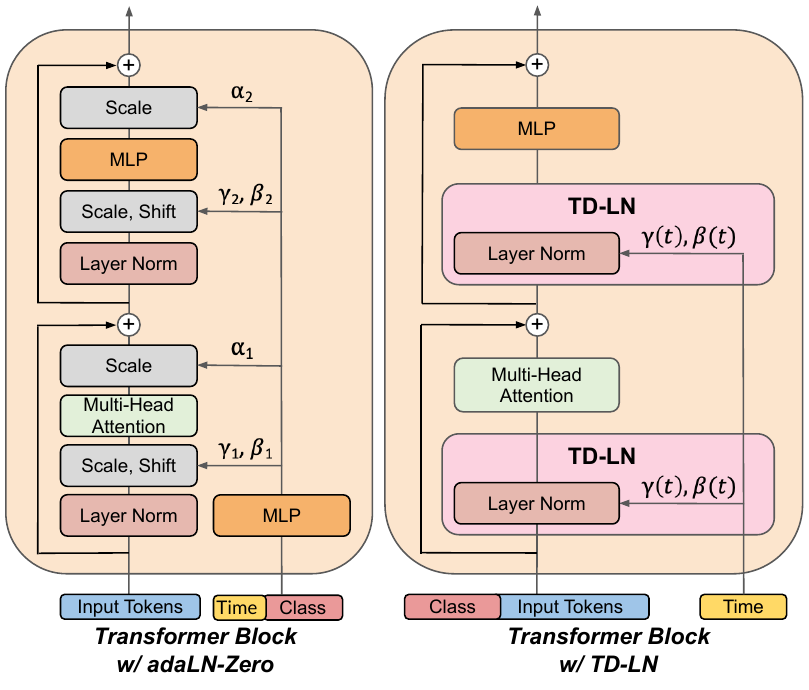}
    \caption{\textbf{Time conditioning mechanisms.} (Left) adaLN-Zero~\cite{peebles2023scalable} learns scale and shift parameters ($\gamma_i$, $\beta_i$, $\alpha_i$, $i=\{1,2\}$) using parameter-heavy MLPs. (Right) The proposed Time-Dependent Layer Normalization (\tdpname) formulates the LN statistics as functions of time ($\gamma(t)$, $\beta(t)$), making it parameter-efficient.}
    \label{fig:td-ln}
\end{wrapfigure}
To understand how time information is utilized in adaLN-Zero, we conducted Principal Component Analysis (PCA) on the learned scale $(\gamma_1, \gamma_2)$ and shift $(\beta_1, \beta_2)$ parameters from a parameter-heavy MLP in adaLN-Zero using a pre-trained DiT-XL/2~\cite{peebles2023scalable} model, as shown in~\figref{fig:adaln}. Intriguingly, we observed that the learned parameters can be largely explained by two principal components, suggesting that a parameter-heavy MLP might be unnecessary and that a simpler function could suffice.
To address the increase in parameters, we introduce Time-Dependent Layer Normalization (\tdpname), a straightforward and lightweight method to inject time into layer normalization. We detail the designs below.


\textbf{adaLN design.}
Building on layer normalization~\cite{ba2016layer}, adaLN additionally learns the scale parameter $\gamma_1$ and shift parameter $\beta_1$ via an MLP from the sum of the embedding vectors of time $t$ and class condition $c$.
Formally, given the input $\bm{x}$ (ignoring the dependency on $t$ for simplicity), we have:
\begin{align}
    \gamma_1, \beta_1 &= \text{MLP}(\text{Embed}(t) + \text{Embed}(c)), \\
    \bm{z} &= \gamma_1 \cdot \text{LN}(\bm{x}, \gamma, \beta) + \beta_1,
\end{align}
where $\gamma_1$ and $\beta_1$ scale and shift the output from the layer normalization $\text{LN}$, the function $\text{Embed}$ generates the embedding vectors for time $t$ and class condition $c$, and $\bm{z}$ is the output.
The $\text{LN}$ has its own learnable affine transform parameters $\gamma$ and $\beta$.
adaLN-Zero~\cite{peebles2023scalable} introduces another scale parameter $\alpha_1$, obtained from the same MLP for zero initialization of a residual block~\cite{goyal2017accurate}.
We note that DiT employs two sets of ($\gamma_1, \beta_1, \alpha_1$) and ($\gamma_2, \beta_2, \alpha_2$) in a Transformer block, as shown in~\figref{fig:td-ln}.

\textbf{\tdpname design.}
In contrast, our proposed method, Time-Dependent Layer Normalization (\tdpname), directly incorporates time $t$ into layer normalization by formulating LN's learnable affine transform parameters $\gamma$ and $\beta$ as functions of $t$.
Motivated by the observation that the learned parameters of adaLN-Zero can be largely explained by two principal components, we propose to model this through the linear interpolation of two learnable parameters $p_1$ and $p_2$. Formally,
\begin{align}
    s(t) &= \text{Sigmoid}(w \cdot t + b), \\
    \gamma(t) &= s(t)\cdot p_1 + (1-s(t)) \cdot p_2,
\end{align}
where $s(t)$ is a transformation of time $t$, $w$ and $b$ are the learnable weight and bias, and $\text{Sigmoid}$ is the sigmoid activation function.
The other affine transform parameter, $\beta(t)$, is formulated similarly with another two parameters $p_3$ and $p_4$.
Consequently, the proposed \tdpname is represented as follows:
\begin{align}
    \bm{z} = \text{LN}(\bm{x},\gamma(t),\beta(t)).
\end{align}
Unlike adaLN, which learns additional re-scaling $\gamma_1$ and re-centering $\beta_1$ variables, \tdpname directly incorporates the time-dependent $\gamma(t)$ and $\beta(t)$ into layer normalization,  eliminating the need for a parameter-heavy MLP.
Furthermore, \tdpname is a versatile mechanism, enabling the injection of time information into both Transformer blocks and ConvNeXt blocks. 
In \modelname, we replace all layer normalizations with the proposed \tdpname, and treat the class condition $c$ as input token for the Transformer blocks.

\subsection{Micro-Level Design}
\label{sec:micro}

In addition to the major architectural modifications discussed earlier, we also explore several micro-level design changes to enhance model performance.


\textbf{Multi-scale loss.}
The proposed Multi-Resolution Network comprises $R$ branches, each dedicated to processing features at a specific resolution, naturally producing multi-scale outputs.
To leverage this, we explore training the network with a multi-scale loss $\mathcal{L}_{multi}$, which is a weighted sum of mean squared error loss at each resolution. Formally, the multi-scale loss is defined as follows:
\begin{equation}
\mathcal{L}_{multi} = \sum_{r=1}^R \alpha_r \cdot \mathbb{E}_{t, \bm{x}_0, c, \bm{\epsilon}_t} \|\text{Downsample}(\bm{\epsilon}_t, r) - \bm{\epsilon}_{\bm{\theta}}(\bm{x}_t, c, r)\|_2^2,
\end{equation}
where $\alpha_r$ is the loss weight for the $r$-th branch, and $\text{Downsample}(\bm{\epsilon}_t, r)$ downsamples the target noise $\bm{\epsilon}_t$ by a factor of $2^{R-r}$ using average pooling (the $R$-th branch, containing no downsampling, is our final output).
We set $\alpha_r = 1 / (2^{R-r}\times2^{R-r})$,  motivated by the prior work~\cite{hoogeboom2023simple} which found that the signal to noise ratio increases by a factor of $k^2$ when the noised input is average-pooled with a $k\times k$ kernel.
Intuitively, our target output (the $R$-th branch) has a  loss weight $\alpha_R=1$, and the loss weights for the intermediate outputs are scaled down quadratically based on the downsampling factor.


\textbf{Gated linear unit.}
In the proposed Multi-Resolution Network, both Transformer and ConvNeXt blocks include an MLP block, consisting of two linear transformations with GeLU activation~\cite{hendrycks2016gaussian} in between.
We also explore replacing the first linear layer with GeGLU~\cite{shazeer2020glu}, an enhanced version of the Gated Linear Unit (GLU)~\cite{dauphin2017language} that has $2\times$ expansion rate.


\subsection{\modelname Model Variants}
\label{sec:meta_arch}
We now introduce the \modelname model variants, scaled appropriately for different model sizes. We present four sizes: \modelname-M (medium, 133M parameters), \modelname-L (large, 284M parameters), \modelname-XL (extra-large, around 500M parameters) and \modelname-G (giant, 1.06B parameters).
Three hyperparameters—$R$ (number of branches), $N$ (number of layers per branch), and $D$ (hidden size per branch)—define each \modelname variant. Specifically, $R$ determines the number of branches in the multi-resolution network. We append 2R or 3R to the model name to indicate whether two or three branches are used. The number of layers $N$ in the multi-resolution network is represented as a tuple of $R$ numbers, where the $r$-th number specifies the number of layers in the $r$-th branch. Similarly, the hidden size $D$ is also a tuple of $R$ numbers.
We follow a straightforward scaling rule: most layers are stacked in the first branch, which is processed by Transformer blocks, while the remaining branches use only half the number of layers of the first branch. Additionally, when the resolution is doubled, the hidden size is reduced by a factor of two.
The model variants details are presented in Tab.~\ref{tab:model_variants} in Sec.~\ref{sec:meta_arch_details} in the Appendix.

\section{Experimental Results}
\label{sec:experiments}

\subsection{Experimental Setup}
\label{sec:experiments-setup}

\noindent\textbf{Datasets.}
We consider class-conditional image generation tasks at $64\times64$, $256\times256$, and $512\times512$ resolutions on ImageNet-1K~\cite{deng2009imagenet}. For images at $64\times64$, we train \modelname on pixel space. For images at $256\times256$ and $512\times512$, following the baselines~\cite{bao2023all, peebles2023scalable}, we utilize an off-the-shelf pre-trained variational autoencoder~\cite{kingma2013auto} from Stable Diffusion~\cite{rombach2022high} to extract the latent representations sized at $32\times32$ and $64\times64$, respectively. Then we train our \modelname to model these latent representations.

\noindent\textbf{Evaluation.}
We measure the model's performance using Fr\'{e}chet Inception Distance (FID)~\cite{heusel2017gans}. 
We report FID on 50K generated samples to measure the image quality (\ie, FID-50K).
To ensure fair comparisons, we follow the same evaluation suite as the baseliness~\cite{dhariwal2021diffusion, peebles2023scalable} to compute the FID scores. We also report Inception Score~\cite{salimans2016improved} and Precision/Recall~\cite{kynkaanniemi2019improved} in~\secref{sec:additional-experiments} as secondary metrics.

\noindent\textbf{Implementation details.}
We use AdamW optimizer~\cite{loshchilov2017decoupled} with a constant learning rate of $2\times 10^{-4}$ for most experiments, except for the $64\times64$ models where we use $3\times 10^{-4}$. A batch size of 1024 is used for most architectures. 
For a fair comparison with DiT~\cite{peebles2023scalable}, we train the $256\times256$ and $512\times512$ models for 1M iterations, and also report results for 500K iterations to compare with U-ViT~\cite{bao2023all}. 
We train the $64\times64$ models for 300K iterations, following the U-ViT protocol. 
\textit{Our training hyperparameters are almost entirely retained from U-ViT~\cite{bao2023all}. We did not tune learning rates, decay/warm-up schedules, Adam $\beta_1$/$\beta_2$ values, or weight decays.} 
Further details on hyperparameters and configurations are provided in Sec.~\ref{sec:setup} in the Appendix.


\subsection{State-of-the-Art Diffusion Models}
\begin{table*}[t!]
\caption{\textbf{Class-conditional image generation on
ImageNet $256\times256$ and ImageNet $512\times512$.} We report training epochs, number of parameters (\#Params), GFLOPs, and FID-50K with and without Classifier-Free Guidance (CFG). Best results are marked in \textbf{bold}. 
}
\label{tab:in1k}
\centering
\subfloat[
\textbf{ImageNet $256\times256$}
\label{tab:in1k256}
]
{
\centering
\begin{minipage}{0.49\linewidth}{\begin{center}
\tablestyle{2.5pt}{1.3}
\scalebox{0.73}{
\begin{tabular}{l|c|cc|cccc}
Model & Epoch & \#Params. & Gflops  & FID (w/o CFG)$\downarrow$ & FID$\downarrow$\\
\shline
ADM-U~\cite{dhariwal2021diffusion} & 396 & 608M & 742 &  7.49 & 3.94 \\
LDM-4 ~\cite{rombach2022high} & 166 & 400M & 104 & 10.56 & 3.60 \\
U-ViT-H/2 ~\cite{bao2023all} & 400 & 501M  & 133 & 6.58 & 2.29 \\
DiT-XL/2~\cite{peebles2023scalable} & 1399 & 675M & 119 & 9.62 & 2.27\\
\modelname-XL/2R (Ours) & 400 & 505M  & 160 & 4.87 & 1.77 \\
\modelname-XL/2R (Ours) & 800 & 505M & 160 & 4.50 & 1.70 \\
\modelname-G/2R (Ours) & 800 & 1.06B & 331 & \textbf{3.56} & \textbf{1.63} \\
\end{tabular}
}
\end{center}}\end{minipage}
}
\subfloat[
\textbf{ImageNet $512\times512$}
\label{tab:in1k512}
]
{
\centering
\begin{minipage}{0.49\linewidth}{\begin{center}
\tablestyle{2.5pt}{1.3}
\scalebox{0.73}{
\begin{tabular}{l|c|cc|ccc}
Model & Epoch & \#Params. & Gflops  & FID (w/o CFG)$\downarrow$ & FID$\downarrow$  \\
\shline
ADM~\cite{dhariwal2021diffusion} & -  & 422M & - & 23.24 & 7.72 \\
ADM-U & 1081 & 731M & 2813 & 9.96 & 3.85 \\
U-ViT-L/4 ~\cite{bao2023all} & 400 & 287M & 77 & - & 4.67 \\
U-ViT-H/4 ~\cite{bao2023all} & 400 & 501M & 133 & 15.70 & 4.05 \\
DiT-XL/2~\cite{peebles2023scalable} & 599 & 675M & 525 & 12.03 & 3.04 \\
\modelname-XL/3R (Ours) & 400 & 525M & 206 & 8.56 & 3.23 \\
\modelname-XL/3R (Ours) & 800 & 525M & 206 & \textbf{7.93} & \textbf{2.89} \\
\end{tabular}
}
\end{center}}\end{minipage}
}
\vspace{-3ex}
\end{table*}


We compare \modelname with state-of-the-art diffusion models on ImageNet $256\times256$ and $512\times512$ in Tab.~\ref{tab:in1k}, and provide more comparisons with other types of generative models in \tabref{tab:full_in1k256} and \tabref{tab:full_in1k512} in the Appendix. 
Results on ImageNet $64\times64$ are reported in~\tabref{tab:full_in1k64} in the Appendix. 
More random samples of the generated images are also presented in Fig.~\ref{fig:supp_512_1} to Fig.~\ref{fig:supp_512_12} in the Appendix.


\noindent\textbf{ImageNet $256\times 256$.} 
From Tab.~\ref{tab:in1k256}, we observe that our \modelname-XL/2R outperforms all previous diffusion-based models and achieves a state-of-the-art FID-50K score of 1.70. Specifically, with a comparable model size and equal or fewer training epochs, our model surpasses previous state-of-the-art transformer-based diffusion models, including U-ViT\cite{bao2023all} (1.77 \vs 2.29 with Classifier-Free Guidance~\cite{ho2022classifier} (CFG) and 4.87 \vs 6.58 without CFG) and DiT~\cite{peebles2023scalable} (1.70 \vs 2.27 with CFG and 4.50 \vs 9.62 without CFG). Our best model, \modelname-G/2R, scales up to the billion-parameter level, setting a new state-of-the-art with an FID of 1.63 with CFG and 3.56 without CFG.


\noindent\textbf{ImageNet $512\times 512$.} 
Our \modelname outperforms all previous diffusion-based models on ImageNet $512 \times 512$ and achieves a state-of-the-art FID-50K score of 2.89 as shown in Tab.~\ref{tab:in1k512}. It is worth noting that, although both Gflops and model sizes are critical for improving performance, as discussed in the DiT paper~\cite{peebles2023scalable}, we still outperform it with only $39.2\%$ of the GFLOPs and $77.8\%$ of the model size, improving the FID-50K from 3.04 to 2.89. 
As transformers and diffusion models have demonstrated good scaling behavior, we believe that further scaling up our \modelname will lead to better performance, which we have left as future work.


\subsection{Alleviating Distortion}
\begin{figure*}
    \centering
    \includegraphics[width=\linewidth]{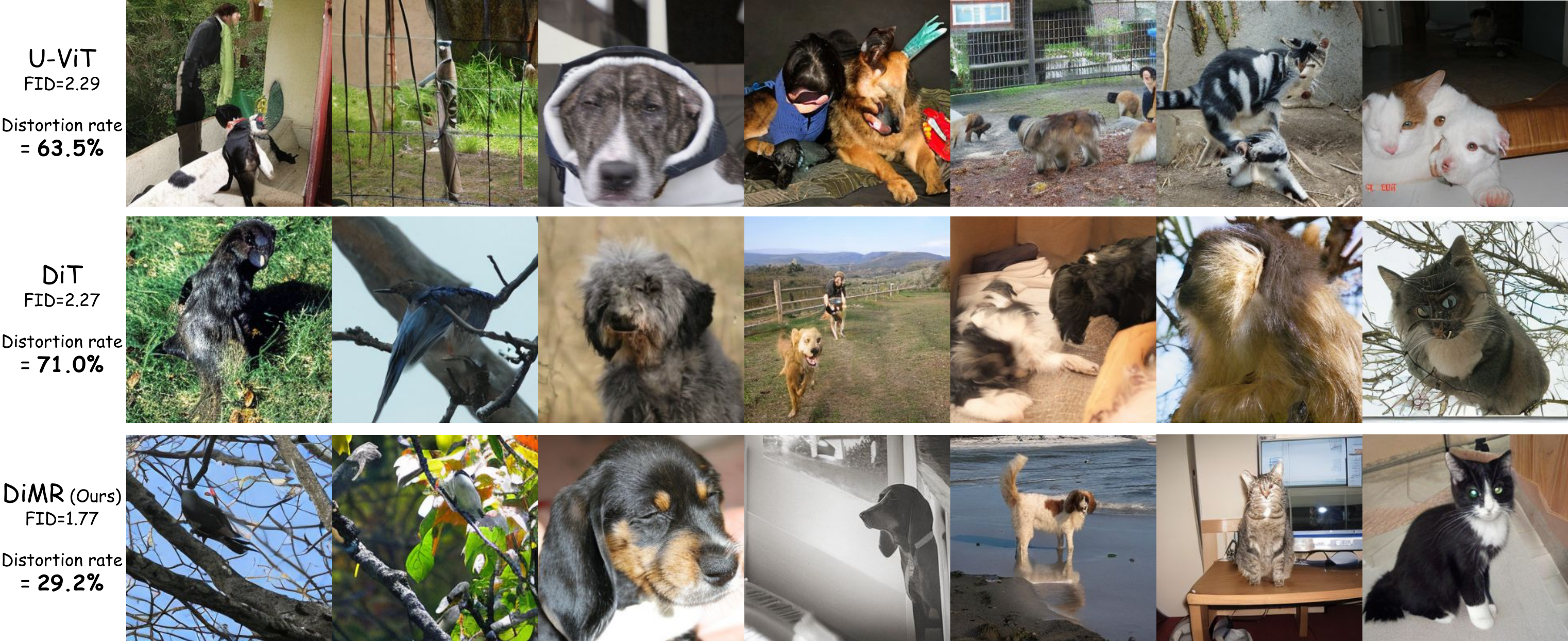}
    \caption{
    \textbf{\modelname alleviates distortions and improves visual fidelity.} 
    In this figure, we randomly visualize the detected low-fidelity images, identified by a pretrained classifier, which are generated by the best models from the baselines and our \modelname. The first column reports both their FID-50K scores and the proportion of distorted images based on human evaluation.
    \modelname demonstrates better generation performance and lower distortion rates than the baselines.
    }
    \label{fig:DiMR-distortion}
\end{figure*}

Transformer-based architectures encounter the challenge of balancing visual fidelity with computational complexity. Despite adopting a small patch size of 2, current models still struggle with distortions. To illustrate the effectiveness of \modelname in alleviating these distortions, we adopt a classifier-based rejection model following previous work~\cite{razavi2019generating}. However, we diverge from previous approaches by \textit{solely} using the rejection model to analyze distorted images, rather than filtering out bad images and computing metrics only on selected `good' images. It is important to note that all metrics in our paper are computed without using the rejection model to ensure fair comparisons.

Specifically, we randomly generate 80K images for each model and utilize a pretrained Vision Transformer classifier~\cite{dosovitskiy2020image} to identify low-fidelity images based on the predicted probabilities. 
Images with a probability below a threshold of 0.2 are considered low-fidelity or potentially distorted.
Fig.~\ref{fig:DiMR-distortion} shows random samples of low-fidelity images detected by the classifier. 
However, we find that not all detected images are distorted; many are classified with low probability due to classifier errors. 
To accurately identify distorted images among those detected by the classifier, we conduct user studies where human evaluators manually assess the images.
Images generated by all three methods are merged and presented, along with their corresponding class labels, to human evaluators, who are instructed to determine whether each image is distorted (\ie, identify low-fidelity images).
Each image is evaluated by five different human evaluators. 
We consider the proportion of distorted images generated by different models, \ie distortion rate.
We compute three distortion rates, one for each model, from each evaluator based on the images they evaluate. 
The final distortion rate for each model is obtained by averaging the rates from all evaluators. 
As reported in Fig.~\ref{fig:DiMR-distortion}, we observe that even among those low-fidelity images, only $29.2\%$ of the images generated by \modelname are distorted, while previous methods yield much higher distortion rates of $63.5\%$ and $71.0\%$.




\begin{table*}[t!]
\tablestyle{3pt}{1.3}
\small
\centering
\caption{\textbf{Ablation study}. Beginning with the baseline, we verify the effectiveness of each component.}
\resizebox{0.9\textwidth}{!}{
\begin{tabular}{cl||cc|ccc|cc}
& Model & AdaLN-Zero~\cite{peebles2023scalable}& \tdpname & Multi-branch & GLU~\cite{shazeer2020glu} & Multi-scale Loss & FID($\downarrow$) & \# Params.  \\
\hline
1 & Baseline (U-ViT-M/4~\cite{bao2023all}) & & & & & & 5.85 & 130.9M \\
2 &  & \checkmark & & & & & 5.44 & 202.4M \\
3 &  & \checkmark & & \checkmark & & & 7.91 & 217.9M \\
4 &  & & \checkmark & \checkmark & & & 5.21 & 154.0M \\
5 &  & & \checkmark & \checkmark & \checkmark & & 4.86 & 132.9M \\
6 & \modelname-M/3R (Ours) & & \checkmark & \checkmark & \checkmark & \checkmark & 3.65 & 132.9M \\
\end{tabular}
}
\label{tab:ablation}
\end{table*}

\subsection{Ablation Studies}
We conduct the primary ablation experiments on ImageNet $64\times 64$, progressively building on the baseline U-ViT-M/4~\cite{bao2023all} to validate the effectiveness of the proposed designs, leading to our final model, \modelname-M/3R, as presented in Tab.~\ref{tab:ablation}.
Additionally, we explore alternative design choices on ImageNet $256 \times 256$ with \modelname-XL/2R, including adopting a pure convolutional architecture, replacing addition with concatenation in feature cascading, and introducing skip connections between branches, as shown in Tab.~\ref{tab:design}.

\textbf{AdaLN-Zero \vs \tdpname.} Since the time token used in U-ViT cannot be adopted for ConvNeXt blocks, we first apply AdaLN-Zero~\cite{peebles2023scalable} to the original U-ViT and our multi-branch network. As observed in row 2 of Tab.~\ref{tab:ablation}, AdaLN-Zero slightly improves the performance of U-ViT from 5.85 to 5.44. However, it does not perform well on ConvNeXt blocks and thus decreases the performance from 5.44 to 7.91 (row 3). Additionally, AdaLN-Zero significantly increases the model size from 130.9M to 202.4M. In contrast, our \tdpname is more flexible and parameter-efficient: it efficiently provides time information to both Transformer blocks and ConvNeXt blocks, improving the FID-50K score from 7.91 to 5.21 (row 4), and also reduces the model size from 217.9M to 154.0M.

\textbf{GLU further reduces model size.} In Tab.~\ref{tab:ablation}, row 5 shows the improvement of GLU compared with the vanilla MLP block. We observe that using GLU slightly improves the performance from 5.21 to 4.86 and further reduces the model size from 154.0M to 132.9M.

\textbf{Multi-scale loss is critical for multi-resolution network.} Training a multi-resolution network presents additional challenges and can result in sub-optimal results. In Tab.~\ref{tab:ablation}, row 6 illustrates that our multi-scale loss significantly enhances the performance, achieving a FID-50K score of 3.65.

\textbf{Multi-branch design improves visual fidelity and alleviates distortions in image generations.} Finally, comparing the multi-branch design in row 6 (incorporating \tdpname, GLU, and multi-scale loss to facilitate training) with the baseline in row 1 reveals a significant improvement in FID-50K, from 5.85 to 3.65, with just a 1.5\% increase in model size (130.9M to 132.9M). Additionally, from Fig.~\ref{fig:DiMR-distortion}, it's evident that the multi-branch design generates images with higher fidelity and less distortion.

\textbf{Transformer is essential for low-resolution processing.} As shown in Table~\ref{tab:conv}, replacing the Transformer blocks in the 1st (lowest-resolution) branch with ConvNeXt blocks results in a \modelname variant that uses only convolutional layers. However, this configuration performs worse compared to combining Transformer blocks with ConvNeXt blocks across different resolutions. This indicates that Transformers are more effective at capturing fine-grained details, while their usage at the lowest resolution maintains a manageable computational cost.

\textbf{Simple addition suffices for multi-resolution feature cascading.} As shown in Table~\ref{tab:cascade}, a straightforward addition operation effectively transfers information from lower-resolution features to higher-resolution features. Replacing addition with concatenation leads to slightly worse results. We also validate the necessity of adding skip-connection between branches. As shown in Table~\ref{tab:skip}, introducing skip-connection not only degrades performance but also complicates the model architecture. Therefore, we adopt a simple upsampling followed by an addition operation for feature cascading.

\begin{table*}[t!]
\caption{\textbf{Design choices.} We empirically experiment with different design choices in model architecture (Tab~\ref{tab:conv}), feature cascade (Tab~\ref{tab:cascade}) and skip-connection (Tab~\ref{tab:skip}). We report FID-50K scores with classifier-free guidance (CFG) after 400 training epochs. 
}
\label{tab:design}
\centering
\subfloat[
\textbf{Pure Conv v.s. Hybrid}
\label{tab:conv}
]
{
\centering
\begin{minipage}{0.33\linewidth}{\begin{center}
\tablestyle{2.5pt}{1.3}
\scalebox{0.73}{
\begin{tabular}{l|c|cc}
Model & 1st branch & FID$\downarrow$ \\ \hline
\multirow{2}{*}{\modelname-XL/2R} & ConvNeXt & 2.09 \\
 & Transformer & 1.77
\end{tabular}
}
\end{center}}\end{minipage}
}
\subfloat[
\textbf{Concatenation v.s. Addition}
\label{tab:cascade}
]
{
\centering
\begin{minipage}{0.33\linewidth}{\begin{center}
\tablestyle{2.5pt}{1.3}
\scalebox{0.73}{
\begin{tabular}{l|c|c}
Model & feature cascade & FID$\downarrow$ \\ \hline
\multirow{2}{*}{\modelname-XL/2R} & Concatenation & 2.06 \\
 & Addition & 1.77
\end{tabular}
}
\end{center}}\end{minipage}
}
\subfloat[
\textbf{w/ v.s. w/o Skip-Connection}
\label{tab:skip}
]
{
\centering
\begin{minipage}{0.33\linewidth}{\begin{center}
\tablestyle{2.5pt}{1.3}
\scalebox{0.73}{
\begin{tabular}{l|c|c}
Model & skip-connection & FID$\downarrow$ \\ \hline
\multirow{2}{*}{\modelname-XL/2R} & \checkmark & 1.96 \\
 &  & 1.77
\end{tabular}
}
\end{center}}\end{minipage}
}
\end{table*}
\section{Conclusion}
In this work, we introduce \modelname, which enhances diffusion models through the Multi-Resolution Network, progressively refining features from low to high resolutions and effectively reducing image distortion. Additionally, \modelname incorporates the proposed parameter-efficient Time-Dependent Layer Normalization (\tdpname), further improving image generation quality. The effectiveness of \modelname has been demonstrated on the popular class-conditional ImageNet generation benchmark, outperforming prior methods and setting new state-of-the-art performance on diffusion-style generative models. 
We hope that \modelname will inspire future designs of both denoising networks and time conditioning mechanisms, paving the way for even more advanced image generation models.

\textbf{Acknowledgement}: We thank Xueqing Deng and Peng Wang for their valuable discussion during Zhanpeng's internship.

\clearpage
{\small
\bibliographystyle{plainnat}
\bibliography{egbib}
}

\clearpage

\appendix
\section*{Appendix}

In the appendix, we provide additional information as listed below:

\begin{itemize}
    \item Sec.~\ref{sec:datasets} provides the dataset information and licenses.
    \item Sec.~\ref{sec:meta_arch_details} provides the \modelname model variants, scaled appropriately for different model sizes.
    \item Sec.~\ref{sec:setup} provides the implementation details of \modelname.
    \item Sec.~\ref{sec:additional-experiments} provides more comparison with other methods for class-conditional image generation on ImageNet $64 \times 64$, ImageNet $256 \times 256$, and ImageNet $512 \times 512$.
    \item Sec.~\ref{sec:additional-pca} provides detailed introduction and more results of the Principal Component Analysis (PCA) on the scale and shift parameters learned in adaLN-Zero.
    \item Sec.~\ref{sec:model-samples} provides more generated image samples by \modelname.
    \item Sec.~\ref{sec:limitations} discusses the limitations of our method.
    \item Sec.~\ref{sec:impacts} discusses the positive societal impacts of our method.
    \item Sec.~\ref{sec:safe} discusses the potential risk of our method and safeguards that will be put in place for responsible release of our models.
\end{itemize}

\section{Datasets Information and Licenses}
\label{sec:datasets}

\textbf{ImageNet:}\quad
The ImageNet~\cite{deng2009imagenet} dataset, containing 1,281,167 training and 50,000 validation images from
1,000 different classes, is a standard benchmark for image classification and class-conditional image generation. For the task of class-conditional image generation, the images are typically resized to a specified size, \eg, $64\times64$, $256\times256$, or $512\times512$.

License: Custom License, non-commercial. \url{https://image-net.org/accessagreement}

Dataset website: \url{https://image-net.org/}

\section{\modelname Model Variants}
\label{sec:meta_arch_details}
We introduce the \modelname model variants, scaled appropriately for different model sizes. We present four sizes: \modelname-M (medium, 132.7M parameters), \modelname-L (large, 284.0M parameters), \modelname-XL (extra-large, around 500M parameters), and \modelname-G (giant, 1.06B parameters)
Three hyperparameters—$R$ (number of branches), $N$ (number of layers per branch), and $D$ (hidden size per branch)—define each \modelname variant. Specifically, $R$ determines the number of branches in the multi-resolution network. We append 2R or 3R to the model name to indicate whether two or three branches are used. The number of layers $N$ in the multi-resolution network is represented as a tuple of $R$ numbers, where the $r$-th number specifies the number of layers in the $r$-th branch. Similarly, the hidden size $D$ is also a tuple of 
$R$ numbers.
We follow a straightforward scaling rule: most layers are stacked in the first branch, which is processed by Transformer blocks, while the remaining branches use only half the number of layers of the first branch. Additionally, when the resolution is doubled, the hidden size is reduced by a factor of two.
The model variants are illustrated in Tab.~\ref{tab:model_variants}.

\begin{table*}[h!]
\tablestyle{3pt}{1.3}
\caption{\textbf{\modelname family}. The specific configuration of a \modelname variant is determined by the hyperparameters $R$ (number of branches), $N$ (number of layers per branch), and $D$ (hidden size per branch).
}
\resizebox{0.75\textwidth}{!}{
\begin{tabular}{l|cc|cccc}
model & input size & latent size & \#branches $R$ & \#layers $N$ & hidden size $D$ & \#params \\
\shline
\modelname-M/3R & $64\times64$ & - & 3 & (15, 8, 8) & (768, 384, 192) & 133M \\
\modelname-L/3R & $64\times64$ & - & 3 & (33, 17, 17) & (768, 384, 192) & 284M \\
\modelname-XL/2R & $256\times256$ & $32\times32$ & 2 & (39, 20) & (960, 480) & 505M \\
\modelname-XL/3R & $512\times512$ & $64\times64$ & 3 & (39, 20, 20) & (960, 480, 240) & 525M \\
\modelname-G/2R & $256\times256$ & $32\times32$ & 2 & (57, 29) & (1152, 576) & 1.06B \\
\end{tabular}
}
\label{tab:model_variants}
\end{table*}

\section{Implementation Details}
\label{sec:setup}
We use the AdamW optimizer~\cite{loshchilov2017decoupled} with a constant learning rate of $2\times 10^{-4}$ for most experiments, except for the $64\times 64$ models where we use $3\times 10^{-4}$. We set the weight decay to 0.03 and the betas to (0.99, 0.99) for all experiments. A batch size of 1024 is used for all architectures.
For a fair comparison with DiT~\cite{peebles2023scalable}, we train the $256\times 256$ and $512\times 512$ models for 1M iterations, and we also report results for 500K iterations to compare with U-ViT~\cite{bao2023all}. We train the $64\times 64$ models for 300K iterations, following the U-ViT protocol. All experiments use 5K steps for warm-up.
We present the detailed experimental setup for all \modelname variants in~\tabref{tab:FGRperCategory}.

\begin{table*}
    \centering
    \caption{\textbf{Experimental setup of \modelname.} Experimental settings for all \modelname variants, including model architectures, training hyperparameters, training costs, and sampler information.}
    \resizebox{\textwidth}{!}{
    \begin{tabular}{lccccc}
    \toprule
    Model & \modelname-M/3R & \modelname-L/3R & \modelname-XL/2R & \modelname-XL/3R & \modelname-G/2R \\
    Parameters & 133M & 284M & 505M & 525M & 1.06B \\
    Image Resolution & $64 \times 64$ & $64 \times 64$ & $256 \times 256$ & $512 \times 512$ & $256 \times 256$ \\
    \midrule
    Latent space & \XSolidBrush & \XSolidBrush & \Checkmark  & \Checkmark & \Checkmark \\
    Latent shape & - & - & $32 \times 32 \times 4$ & $64 \times 64 \times 4 $ & $32 \times 32 \times 4$ \\ 
    Image decoder  & - & - & sd-vae-ft-ema & sd-vae-ft-ema & sd-vae-ft-ema \\
    \midrule
    Number of branches $R$ & 3 & 3 & 2 & 3 & 2 \\
    Blocks in 1st branch & Transformer & Transformer & Transformer & Transformer & Transformer \\
    \qquad \# Layers & 15 & 33 & 39 & 39 & 57 \\
    \qquad \# Dimensions & 768 & 768 & 960 & 960 & 1152 \\
    \qquad \# Heads & 12 & 12 & 16 & 16 & 16 \\
    \qquad Resolution & $16\times 16$ & $16\times 16$ & $16\times 16$ & $16\times 16$ & $16\times 16$ \\
    \qquad Loss Coeffs. & $ 1/16 $ & $ 1/16 $ & $ 1/4 $ & $ 1/16 $ & $ 1/4 $ \\
    Blocks in 2nd branch & ConvNeXt & ConvNeXt & ConvNeXt & ConvNeXt & ConvNeXt \\
    \qquad \# Layers & 8 & 17 & 20 & 20 & 29 \\
    \qquad \# Dimensions & 384 & 384 & 480 & 480 & 576 \\
    \qquad Kernel size & $7\times 7$ & $7\times 7$ & $7\times 7$ & $7\times 7$ & $7\times 7$ \\
    \qquad Resolution & $32\times 32$ & $32\times 32$ & $32\times 32$ & $32\times 32$ & $32\times 32$ \\
    \qquad Loss Coeffs. & $ 1/4 $ & $ 1/4 $ & $ 1 $ & $ 1/4 $ & $ 1 $ \\
    Blocks in 3rd branch & ConvNeXt & ConvNeXt & - & ConvNeXt & - \\
    \qquad \# Layers & 8 & 17 & - & 20 & - \\
    \qquad \# Dimensions & 192 & 192 & - & 240 & - \\
    \qquad Kernel size & $7\times 7$ & $7\times 7$ & - & $7\times 7$ & - \\
    \qquad Resolution & $64\times 64$ & $64\times 64$ & - & $64\times 64$ & - \\
    \qquad Loss Coeffs. & $ 1 $ & $ 1 $ & - & $ 1 $ & -\\
    \midrule
    Batch size & 1024 & 1024 & 1024 & 1024 & 1024 \\
    Training iterations & 300K & 300K & 1M & 1M & 1M \\
    Warm-up steps & 5K & 5K & 5K & 5K & 5K \\
    Optimizer & AdamW & AdamW & AdamW & AdamW & AdamW \\
    learning rate & $3\times 10^{-4}$ & $3\times 10^{-4}$ & $2\times 10^{-4}$ & $2\times 10^{-4}$ & $2\times 10^{-4}$ \\
    Weight decay & $ 0.03 $ & $ 0.03 $ & $ 0.03 $ & $ 0.03 $ & $ 0.03 $ \\
    Betas & $ (0.99, 0.99) $ & $ (0.99, 0.99) $ & $ (0.99, 0.99) $ & $ (0.99, 0.99) $ & $ (0.99, 0.99) $ \\
    Training devices & 8 A100 & 16 A100 & 16 A100 & 16 A100 & 32 A100 \\
    Training time & 62 hours & 80 hours & 172 hours & 288 hours & 400 hours \\
    \midrule
    Sampler & DPM-Solver & DPM-Solver & DPM-Solver & DPM-Solver & DPM-Solver \\
    Sampling steps & 50 & 50 & 250 & 250 & 250 \\
    \bottomrule
    \end{tabular} 
    }
    \label{tab:FGRperCategory}
\end{table*}

\section{Additional Experimental Results}
\label{sec:additional-experiments}
\begin{table*}[t!]
\small
\centering
\caption{\textbf{Class-conditional image generation on ImageNet $64 \times 64$ (w/o classifier-free guidance)}. Metrics include Fréchet Inception Distance (FID), Inception Score (IS), Precision, and Recall, where ``$\downarrow$'' or ``$\uparrow$'' indicate whether lower or higher values are better, respectively. ``Type'': the type of the generative model. ``Epoch'': the number of epochs trained on ImageNet~\cite{deng2009imagenet}. ``\#Params'': the number of parameters in the model. ``\#Gflops'': the computational cost. ``Diff.'': Diffusion models. 
}
\resizebox{1\textwidth}{!}{
\begin{tabular}{l|c|c|cc|cccc}
Model & Type & Epoch & \#Params. & Gflops & FID($\downarrow$) & IS($\uparrow$) & Precision($\uparrow$) & Recall($\uparrow$) \\
\shline
U-ViT-M/4~\cite{bao2023all} & Diff. & 240 & 131M & 35 & 5.85 & 33.71 & 0.69 & 0.61 \\
U-ViT-L/4~\cite{bao2023all} & Diff. & 240 & 287M & 77 & 4.26 & 40.66 & 0.71 & 0.62 \\
\modelname-M/3R (Ours) & Diff. & 240 & 133M & 54 & 3.65 & 42.41 & 0.74 & 0.59 \\
\modelname-L/3R (Ours) & Diff. & 240 & 284M & 111 & 2.21 & 55.73 & 0.75 & 0.60 \\

\end{tabular}}
\label{tab:full_in1k64}
\end{table*}


\begin{table*}[t!]
\small
\centering
\caption{\textbf{Class-conditional image generation on ImageNet $256 \times 256$ (with classifier-free guidance)}. Metrics include Fréchet Inception Distance (FID), Inception Score (IS), Precision, and Recall, where ``$\downarrow$'' or ``$\uparrow$'' indicate whether lower or higher values are better, respectively. 
We report results of GAN-based models (GAN), BERT-style masked-prediction models (Mask.), autoregressive models (AR),  visual autoregressive models (VAR), and diffusion based models (Diff.).
``Type'': the type of the generative model. ``Epoch'': the number of epochs trained on ImageNet~\cite{deng2009imagenet}. ``\#Params'': the number of parameters in the model. ``\#Gflops'': the computational cost. ``-re'': the models utilize rejection sampling. ``Mask. + Diff.": the models using masked-prediction to improve diffusion models.
}
\resizebox{1\textwidth}{!}{
\begin{tabular}{l|c|c|cc|cccc}
Model & Type & Epoch & \#Params. & Gflops & FID($\downarrow$) & IS($\uparrow$) & Precision($\uparrow$) & Recall($\uparrow$) \\
\shline
BigGAN~\cite{brock2018large} & GAN & - & 112M & - & 6.95 & 224.5 & 0.89 & 0.38 \\
GigaGAN~\cite{kang2023scaling} & GAN & - & 569M & - & 3.45 & 225.5 & 0.84 & 0.61 \\
\hline
MaskGIT~\cite{chang2022maskgit} & Mask. & 300 & 227M & - & 6.18 & 182.1 & 0.80 & 0.51 \\
MaskGIT-re~\cite{chang2022maskgit} & Mask. & 300 & 227M & 300 & 4.02 & 355.6 & - & - \\
RCG~\cite{li2023self} & Mask. & 200 & 502M & - & 3.49 & 215.5 & - & - \\
TiTok-S-128~\cite{yu2024image} & Mask. & 800 & 287M & - & 1.97 & 281.8 & - & - \\
MDT-G~\cite{gao2023masked} & Mask. + Diff. & 1299 & 676M & 119 & 1.79 & 283.0 & 0.81 & 0.61 \\
MDTv2-G~\cite{gao2023mdtv2} & Mask. + Diff. & 919 & 675M & 119 & 1.58 & 314.7 & 0.79 & 0.65 \\
MaskBit~\cite{mark2024maskbit} & Mask. & 1080 & 305M & - & 1.52 & 328.6 & - & - \\
\hline
VQGAN~\cite{esser2021taming} & AR & 100 & 1.4B & - & 15.78 & 74.3 & - & - \\ 
VQGAN-re~\cite{esser2021taming} & AR & 100 & 1.4B & - & 5.20 & 280.3 & - & - \\ 
ViTVQ~\cite{yu2021vector} & AR & 100 & 1.7B & - & 4.17 & 175.1 & - & - \\ 
ViTVQ-re~\cite{yu2021vector} & AR & 100 & 1.7B & - & 3.04 & 227.4 & - & - \\ 
RQTran~\cite{lee2022autoregressive} & AR & 50 & 3.8B & - & 7.55 & 134.0 & - & - \\ 
RQTran-re~\cite{lee2022autoregressive} & AR & 50 & 3.8B & - & 3.80 & 323.7 & - & - \\ 
\hline
VAR-${d}16$~\cite{tian2024visual} & VAR & 200 & 310M & - & 3.60 & 257.5 & 0.85 & 0.48 \\
VAR-${d}20$~\cite{tian2024visual} & VAR & 250 & 600M & - & 2.95 & 306.1 & 0.84 & 0.53 \\
VAR-${d}24$~\cite{tian2024visual} & VAR & 350 & 1.0B & - & 2.33 & 320.1 & 0.82 & 0.57 \\
VAR-${d}30$~\cite{tian2024visual} & VAR & 350 & 2.0B & - & 1.97 & 334.7 & 0.81 & 0.61 \\
VAR-${d}30$-re~\cite{tian2024visual} & VAR & 350 & 2.0B & - & 1.80 & 356.4 & 0.83 & 0.57 \\
\hline
ADM-G~\cite{dhariwal2021diffusion} & Diff. & 396 & 554M & - & 4.59 & 186.7 & 0.82 & 0.52 \\
ADM-G, ADM-U~\cite{dhariwal2021diffusion} & Diff. & 208 & 608M & 742 & 3.94 & 215.8 & 0.83 & 0.53 \\
CDM~\cite{ho2022cascaded} & Diff. & 2158 & - & - & 4.88 & 158.7 & - & - \\
LDM-4~\cite{rombach2022high} & Diff. & 166 & 400M & - & 3.60 & 247.7 & - & - \\
DiT-L/2~\cite{peebles2023scalable} & Diff. & 1399 & 458M & 81 & 5.02 & 167.2 & 0.75 & 0.57 \\
DiT-XL/2~\cite{peebles2023scalable} & Diff. & 1399 & 675M & 119 & 2.27 & 278.2 & 0.83 & 0.57 \\
U-ViT-L/2~\cite{bao2023all} & Diff. & 240 & 287M & 77 & 3.40 & 219.9 & 0.83 & 0.52 \\
U-ViT-H/2~\cite{bao2023all} & Diff. & 400 & 501M & 133 & 2.29 & 263.9 & 0.82 & 0.57 \\
D\scriptsize{IFFU}\small{SSM}-XL~\cite{yan2023diffusion} & Diff. & 515 & 673M & 280 & 2.28 & 259.1 & 0.86 & 0.56 \\
SiT-XL~\cite{ma2024sit} & Diff. & 1399 & 675M & 119 & 2.06 & 270.3 & 0.82 & 0.59 \\
DiffiT~\cite{hatamizadeh2023diffit} & Diff. & 400 & 561M & 114 & 1.73 & 276.5 & 0.80 & 0.62 \\
\modelname-XL/2R (Ours) & Diff. & 400 & 505M & 160 & 1.77 & 285.7 & 0.79 & 0.62 \\
\modelname-XL/2R (Ours) & Diff. & 800 & 505M & 160 & 1.70 & 289.0 & 0.79 & 0.63 \\
\modelname-G/2R (Ours) & Diff. & 800 & 1.1B & 331 & 1.63 & 292.5 & 0.79 & 0.63 \\
\end{tabular}}
\label{tab:full_in1k256}
\end{table*}


\begin{table*}[t!]
\small
\centering
\caption{\textbf{Class-conditional image generation on ImageNet $512 \times 512$ (with classifier-free guidance)}. Metrics include Fréchet Inception Distance (FID), Inception Score (IS), Precision, and Recall, where ``$\downarrow$'' or ``$\uparrow$'' indicate whether lower or higher values are better, respectively. 
We report results of GAN-based models (GAN), BERT-style masked-prediction models (Mask.), autoregressive models (AR),  visual autoregressive models (VAR), and diffusion based models (Diff.).
``Type'': the type of the generative model. ``Epoch'': the number of epochs trained on ImageNet~\cite{deng2009imagenet}. ``\#Params'': the number of parameters in the model. ``\#Gflops'': the computational cost.
}
\resizebox{1\textwidth}{!}{
\begin{tabular}{l|c|c|cc|cccc}
Model & Type & Epoch & \#Params. & Gflops  & FID($\downarrow$) & IS($\uparrow$) & Precision($\uparrow$) & Recall($\uparrow$) \\
\shline
BigGAN~\cite{brock2018large} & GAN & - & 158M & - & 8.43 & 177.9 & 0.88 & 0.29 \\
\hline
MaskGIT~\cite{chang2022maskgit} & Mask. & 300 & 227M & - & 7.32 & 156.0 & 0.78 & 0.50 \\
MaskGIT-re~\cite{chang2022maskgit} & Mask. & 300 & 227M & - & 4.46 & 342.0 & - & - \\
\hline
VAR-${d}36$-s~\cite{tian2024visual} & VAR & 350 & 2.35B & - & 2.63 & 303.2 & - & - \\
\hline
ADM-G~\cite{dhariwal2021diffusion} & Diff. & - & 422M & - & 7.72 & 172.7 & 0.87 & 0.42 \\
ADM-G, ADM-U~\cite{dhariwal2021diffusion} & Diff. & 1081 & 731M & 2813 & 3.85 & 221.7 & 0.84 & 0.53 \\
DiT-XL/2~\cite{peebles2023scalable} & Diff. & 599 & 675M & 525 & 3.04 & 240.8 & 0.84 & 0.54 \\
U-ViT-L/4~\cite{bao2023all} & Diff. & 400 & 287M & 77 & 4.67 & 213.3 & 0.87 & 0.45 \\
U-ViT-H/4~\cite{bao2023all} & Diff. & 400 & 501M & 133 & 4.05 & 263.8 & 0.84 & 0.48 \\
D\scriptsize{IFFU}\small{SSM}-XL~\cite{yan2023diffusion} & Diff. & 236 & 673M & 1066 & 3.41 & 255.0 & 0.85 & 0.49 \\
DiffiT~\cite{hatamizadeh2023diffit} & Diff. & 800 & 561M & - & 2.67 & 252.1 & 0.83 & 0.55 \\
\modelname-XL/3R (Ours) & Diff. & 400 & 525M & 206 & 3.23 & 285.1 & 0.82 & 0.54 \\
\modelname-XL/3R (Ours) & Diff. & 800 & 525M & 206 & 2.89 & 289.8 & 0.83 & 0.55 \\
\end{tabular}}
\label{tab:full_in1k512}
\end{table*}
We present the full results of the proposed \modelname compared to other methods on ImageNet~\cite{deng2009imagenet} in terms of Fréchet Inception Distance (FID)~\cite{heusel2017gans}, Inception Score (IS)~\cite{salimans2016improved}, and Precision/Recall~\cite{kynkaanniemi2019improved}. The comparisons are made on class-conditional image generation without classifier-free guidance on ImageNet $64 \times 64$ in~\tabref{tab:full_in1k64}, and with classifier-free guidance~\cite{ho2022classifier} on ImageNet $256 \times 256$ in~\tabref{tab:full_in1k256} and ImageNet $512 \times 512$ in~\tabref{tab:full_in1k512}.

\noindent\textbf{ImageNet $64 \times 64$.} We follow the exact experimental setup of U-ViT~\cite{bao2023all} for class-conditional image generation on ImageNet $64 \times 64$ without classifier-free guidance to verify the effectiveness of our proposed backbone. Therefore, we focus solely on comparing against U-ViT on this benchmark. As shown in~\tabref{tab:full_in1k64}, when both are trained for 240 epochs, the proposed \modelname-M/3R with 133M parameters achieves an FID of 3.65 and an IS of 42.41, improving upon the counterpart U-ViT-M/4 with 131M parameters by 2.20 in FID and 8.70 in IS. For the larger model, \modelname-L/3R with 284M parameters outperforms U-ViT-L/4 with 287M parameters by 2.05 in FID and 15.07 in IS. These consistent and significant improvements demonstrate the capability of the proposed Multi-Resolution Network and \tdpname in enhancing diffusion models to generate high-fidelity images.

\noindent\textbf{ImageNet $256 \times 256$.} We compare \modelname with state-of-the-art generative models on ImageNet $256 \times 256$ with classifier-free guidance in Tab.~\ref{tab:full_in1k256}. Compared to U-ViT~\cite{bao2023all} in a fair setting, our \modelname-XL/2R with 505M parameters, trained for 400 epochs, significantly outperforms U-ViT-H/2 with 501M parameters, also trained for 400 epochs, by 0.52 in FID and 21.8 in IS. In comparison with the recently popular diffusion model DiT~\cite{peebles2023scalable}, \modelname-XL/2R trained for 800 epochs consistently outperforms DiT-L/2 with 458M parameters by 3.32 in FID and 121.8 in IS. \modelname-XL/2R even surpasses the larger variant DiT-XL/2 with 675M parameters by 0.57 in FID and 11.8 in IS. Notably, DiT models require training for 1399 epochs, while \modelname-XL/2R achieves superior performance with only 800 epochs. Furthermore, scaling up to \modelname-G/2R sets a new state-of-the-art for ImageNet $256 \times 256$ image generation, achieving an FID of 1.63 and an IS of 292.5. 

\noindent\textbf{ImageNet $512 \times 512$.} We compare \modelname with state-of-the-art generative models on ImageNet $512 \times 512$ with classifier-free guidance in Tab.~\ref{tab:full_in1k512}. Under the same training setting for 400 epochs, \modelname-XL/3R with 525M parameters outperforms U-ViT-H/4 with 501M parameters by 0.82 in FID and 21.3 in IS. Compared to DiT-XL/2 with 675M parameters, \modelname-XL/3R shows slight improvements of 0.15 in FID and 49.0 in IS. Overall, \modelname-XL/3R demonstrates performance comparable to other state-of-the-art generative models for ImageNet $512 \times 512$ image generation.

\section{Additional PCA of Learned Scale and Shift Parameters in adaLN-Zero}
\label{sec:additional-pca}
\begin{figure}[htbp]
  \centering
  \begin{minipage}[t]{\linewidth}  
      \centering
      \label{fig:pca_ratio_6}
      \includegraphics[width=\linewidth]{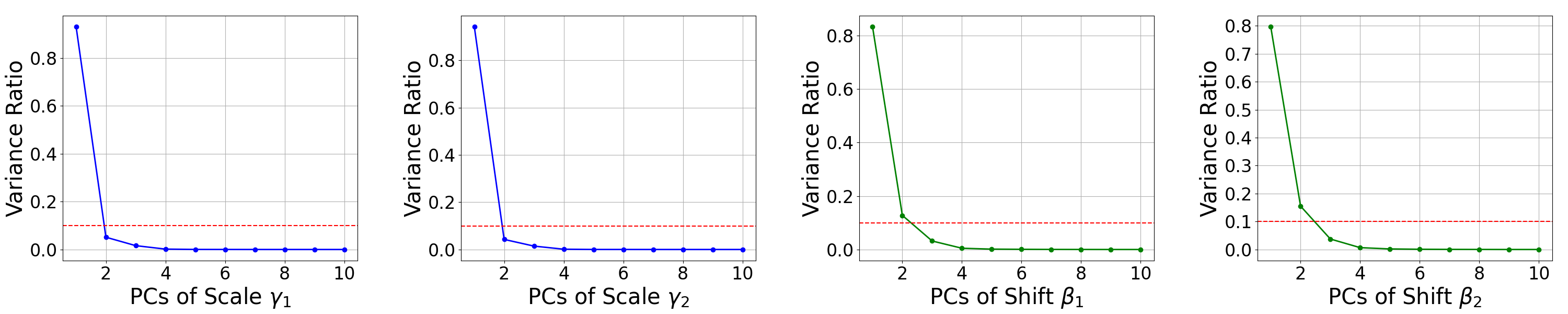}
  \end{minipage}
  \begin{minipage}[t]{\linewidth}
      \centering
      \label{fig:pca_ratio_13}
      \includegraphics[width=\linewidth]{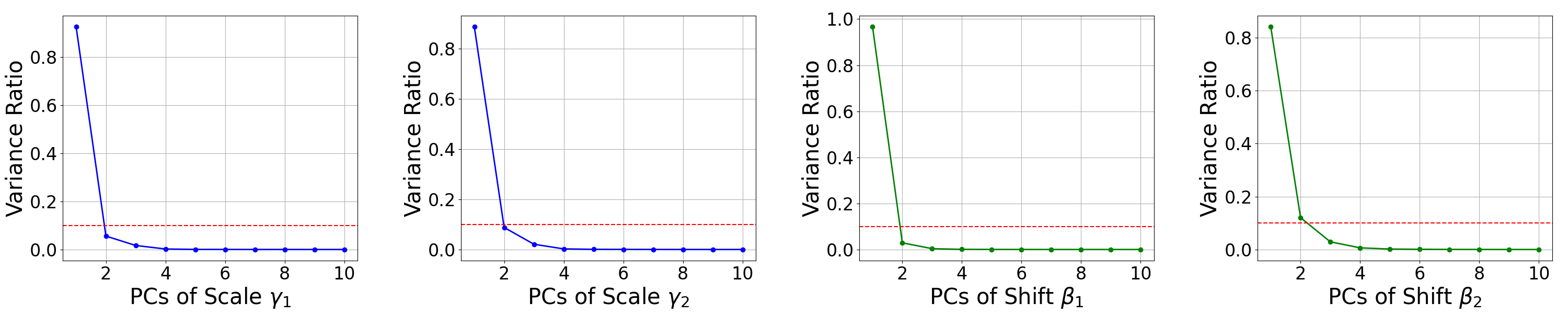}
  \end{minipage}
  \begin{minipage}[t]{\linewidth}
      \centering
      \label{fig:pca_ratio_20}
      \includegraphics[width=\linewidth]{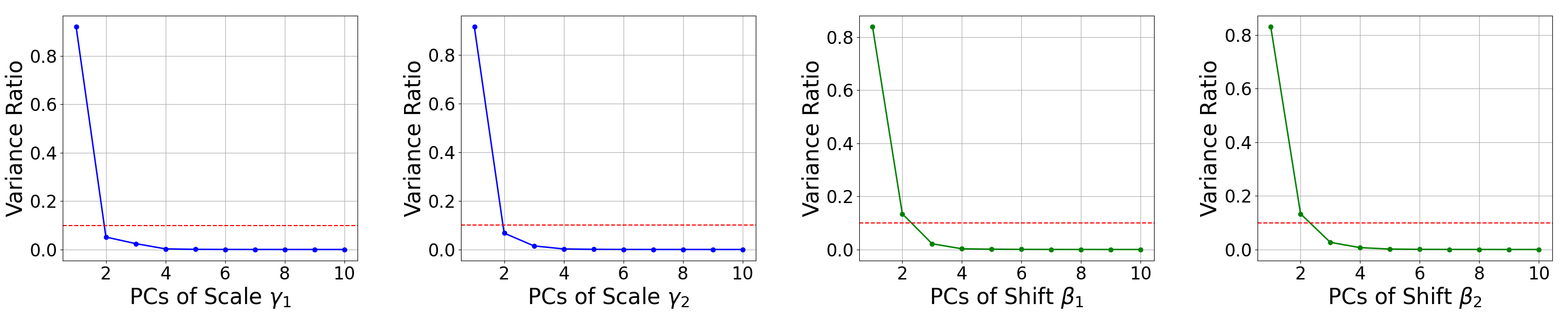}
  \end{minipage}
  \begin{minipage}[t]{\linewidth}
      \centering
      \label{fig:pca_ratio_26}
      \includegraphics[width=\linewidth]{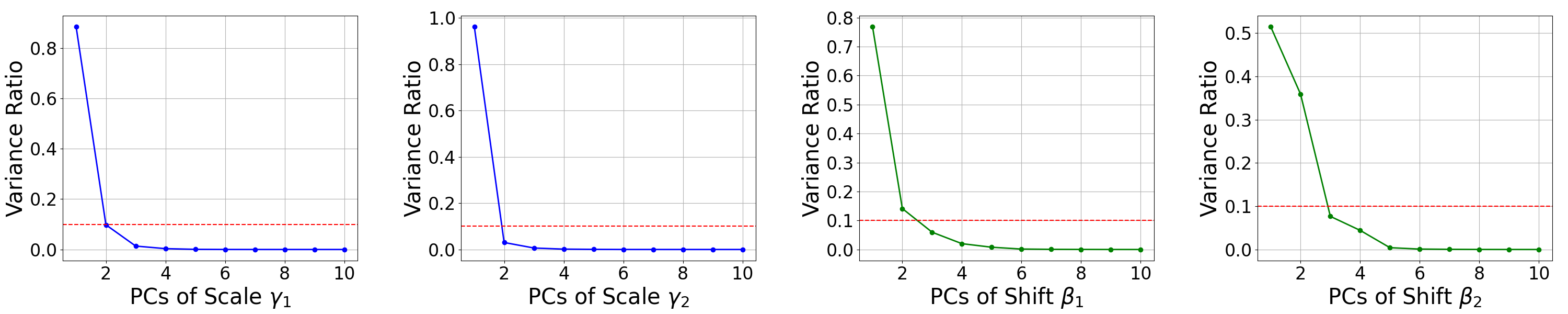}
  \end{minipage}
  \caption{\textbf{Principal Component Analysis (PCA) of learned scale and shift parameters in adaLN-Zero~\cite{peebles2023scalable}.} We conduct PCA on the learned scale ($\gamma_1$, $\gamma_2$) and shift ($\beta_1$, $\beta_2$) parameters obtained from a parameter-heavy MLP in adaLN-Zero using a pre-trained DiT-XL/2~\cite{peebles2023scalable} model with 28 layers in depth. Each row of the figure presents the analysis result at different depths, from top to bottom: 7, 14, 21, 28. The vertical axis represents the explained variance ratio of the corresponding Principal Components (PCs). Our observations reveal that the learned parameters can be largely explained by two principal components, suggesting the potential to approximate them by a simpler function.}
  \label{fig:additional-pca}
\end{figure}

We conduct PCA on the learned scale ($\gamma_1$, $\gamma_2$) and shift ($\beta_1$, $\beta_2$) parameters obtained from a parameter-heavy MLP in adaLN-Zero using a pre-trained DiT-XL/2~\cite{peebles2023scalable} model with 28 layers in depth. To conduct the analysis, we utilize the pre-trained DiT-XL/2 to generate images and collected the scale and shift parameters (tensors) produced by the MLP at different layers along the sampling steps. PCA is then performed on the collected tensors at each layer separately. The results are presented in~\figref{fig:additional-pca}, where each row of the figure displays the analysis result at different depths, from top to bottom: 7, 14, 21, 28. The vertical axis represents the explained variance ratio of the corresponding Principal Components (PCs). As observed, in most cases, the most important principal component can explain most of the variance, while starting from the 3rd principal component, it usually only accounts for less than 5\% of the variance. Our observations reveal that the learned parameters, regardless of whether produced by an MLP at a shallower layer or a deeper layer, can be largely explained by two principal components, suggesting the potential to approximate them by a simpler function, \tdpname, where the linear interpolation of two learnable parameters is learned as introduced in Sec.~\ref{sec:tdln}.

\section{Model Samples}
\label{sec:model-samples}
\begin{figure*}
    \centering
    \includegraphics[width=\linewidth]{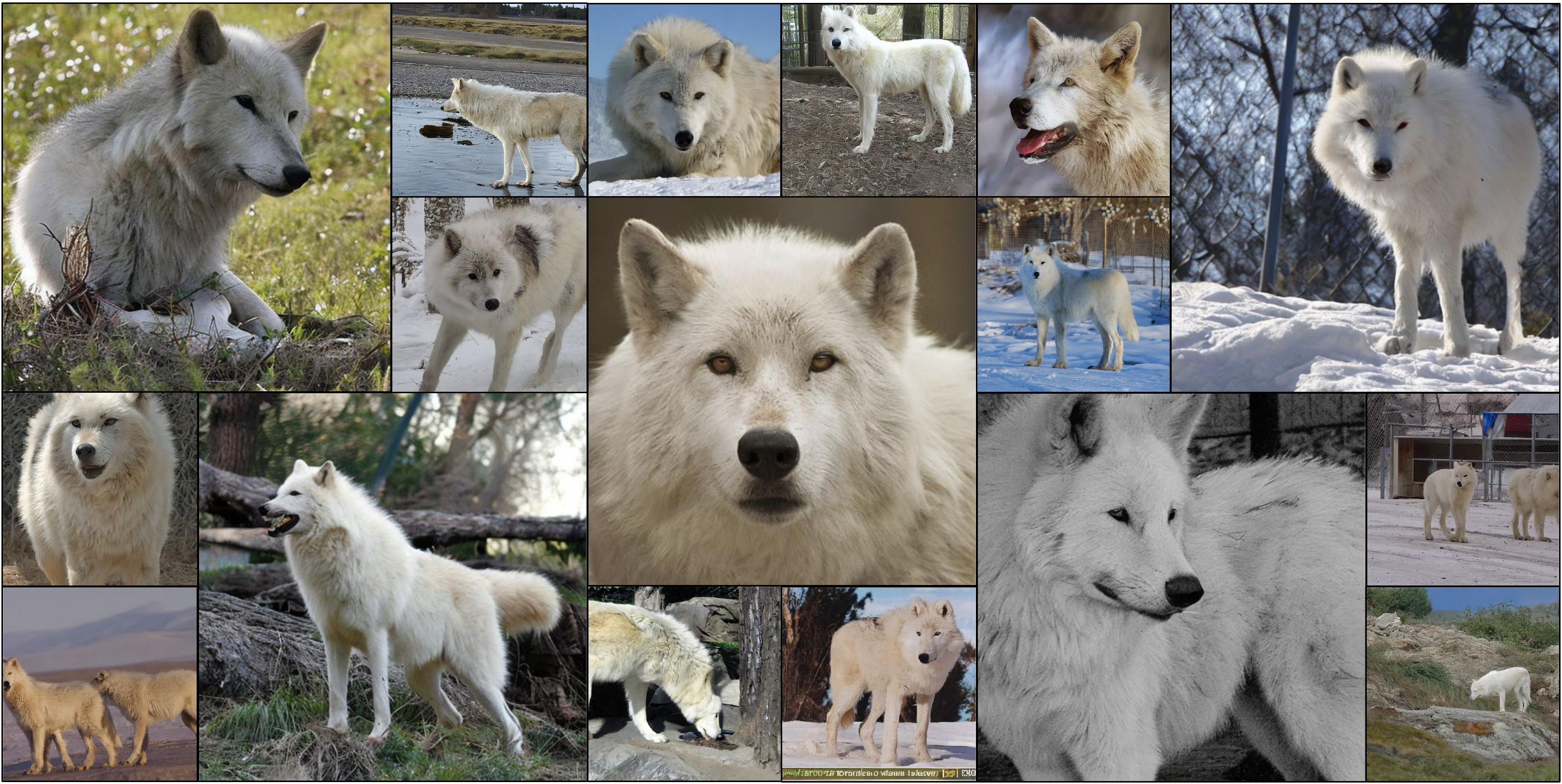}
    \caption{
    \textbf{Uncurated $512 \times 512$ \modelname samples.} Class label = `arctic wolf' (270)}
    \vspace{-3ex}
    \label{fig:supp_512_1}
\end{figure*}

\begin{figure*}
    \centering
    \includegraphics[width=\linewidth]{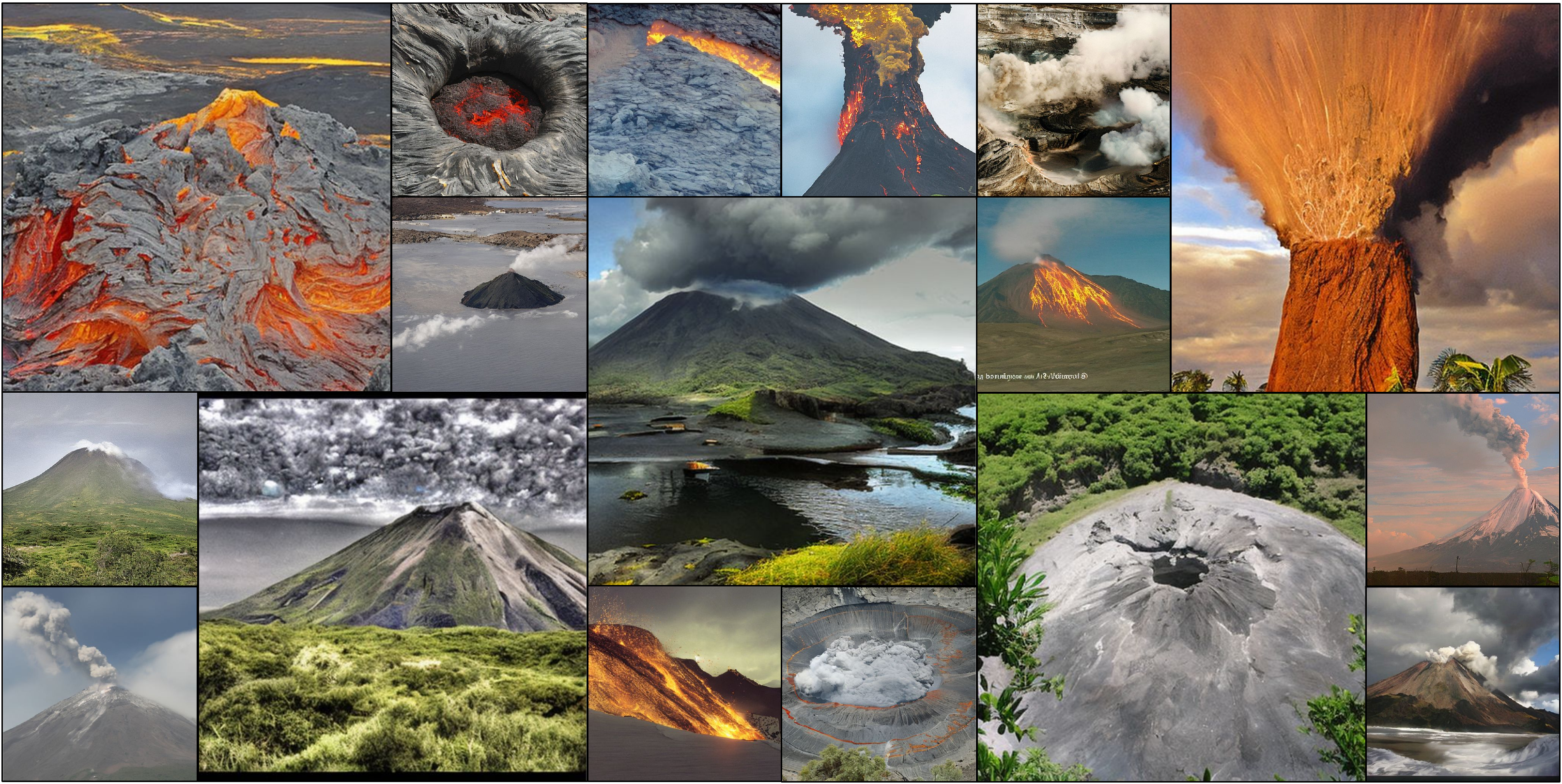}
    \caption{
    \textbf{Uncurated $512 \times 512$ \modelname samples.} Class label = `volcano' (980)}
    \vspace{-3ex}
    \label{fig:supp_512_2}
\end{figure*}

\begin{figure*}
    \centering
    \includegraphics[width=\linewidth]{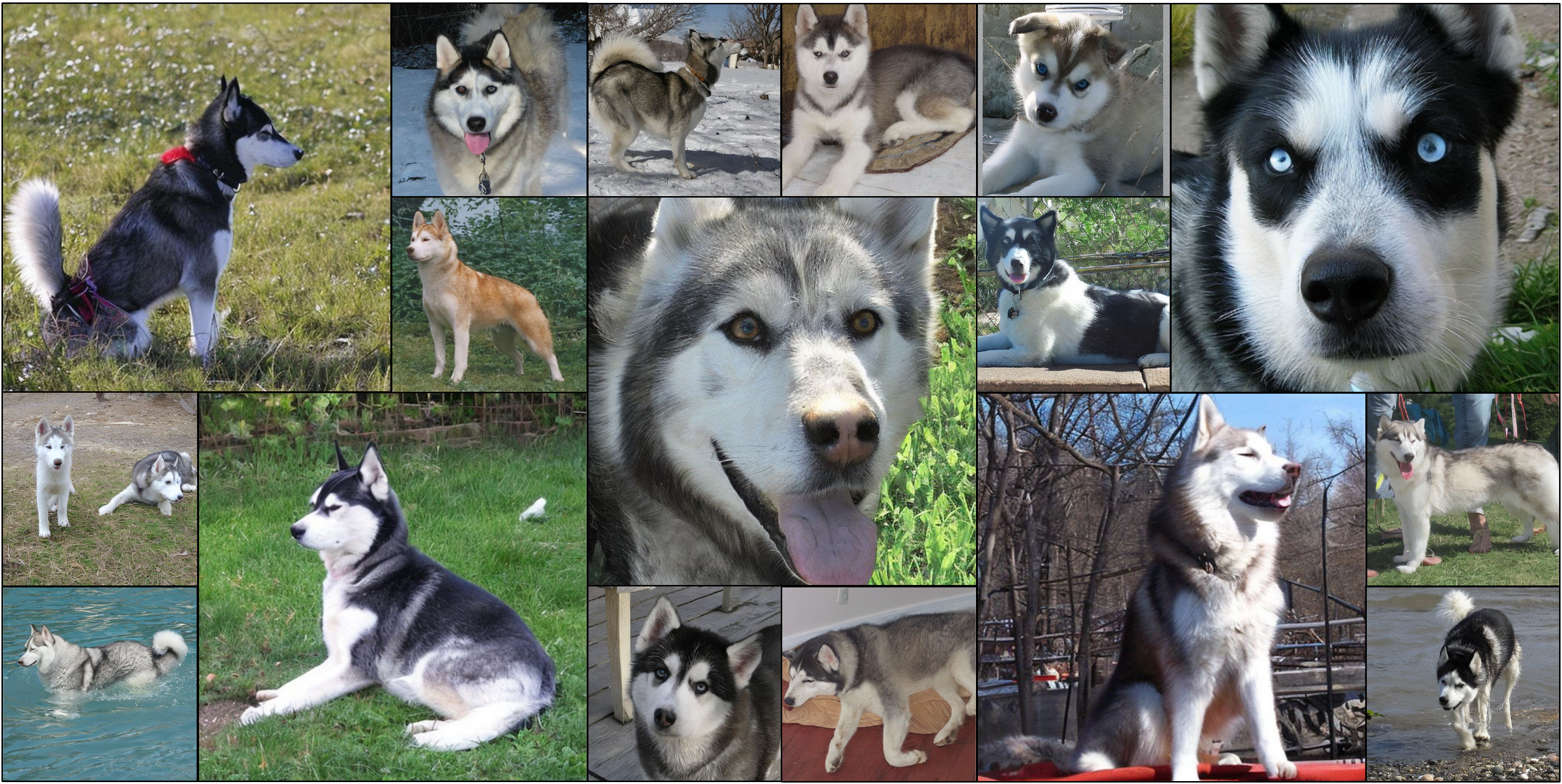}
    \caption{
    \textbf{Uncurated $512 \times 512$ \modelname samples.} Class label = `husky' (250)}
    \vspace{-3ex}
    \label{fig:supp_512_3}
\end{figure*}

\begin{figure*}
    \centering
    \includegraphics[width=\linewidth]{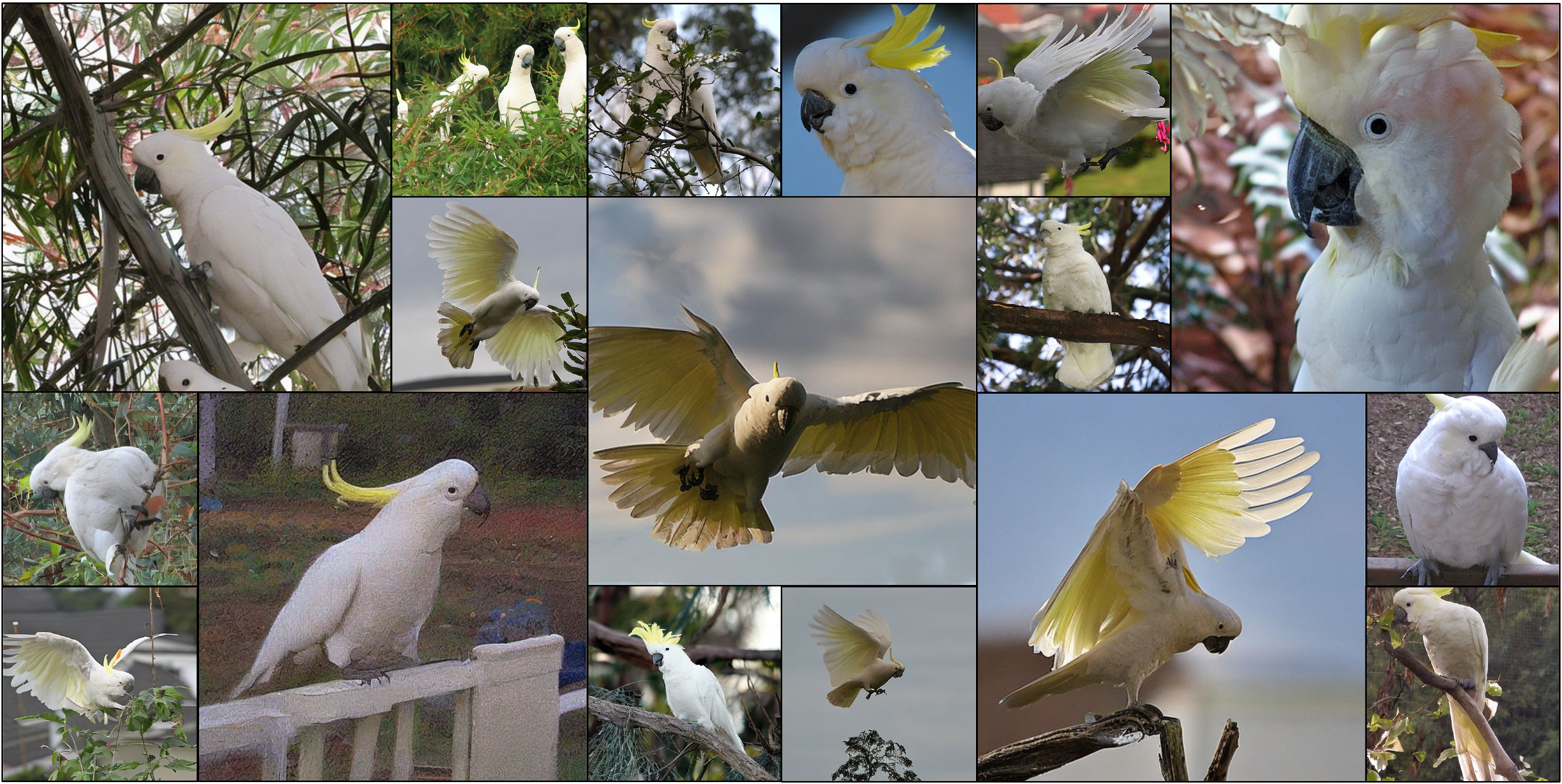}
    \caption{
    \textbf{Uncurated $512 \times 512$ \modelname samples.} Class label = `sulphur-crested cockatoo' (89)}
    \vspace{-3ex}
    \label{fig:supp_512_4}
\end{figure*}

\begin{figure*}
    \centering
    \includegraphics[width=\linewidth]{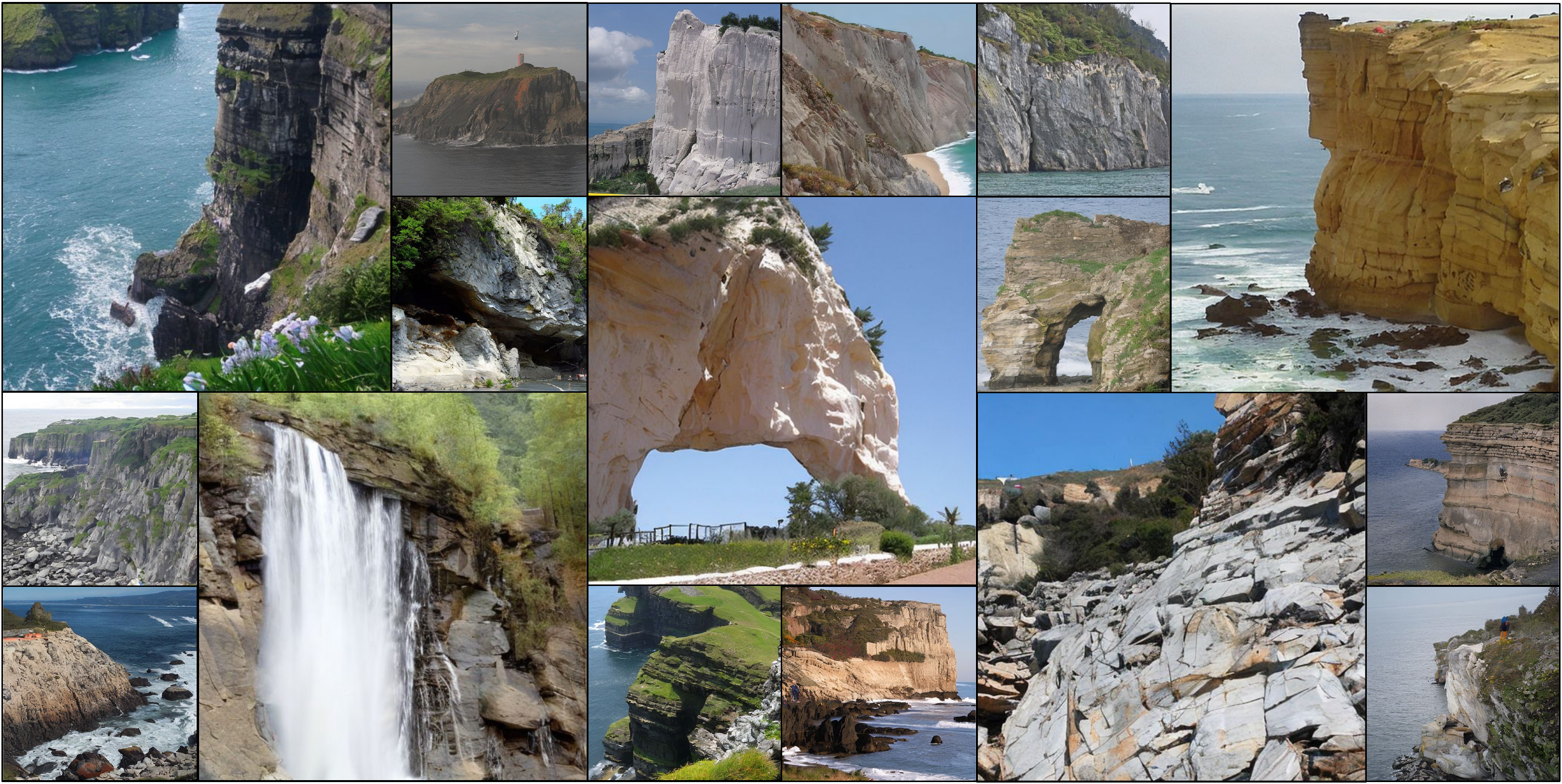}
    \caption{
    \textbf{Uncurated $512 \times 512$ \modelname samples.} Class label = `cliff drop-off' (972)}
    \vspace{-3ex}
    \label{fig:supp_512_5}
\end{figure*}

\begin{figure*}
    \centering
    \includegraphics[width=\linewidth]{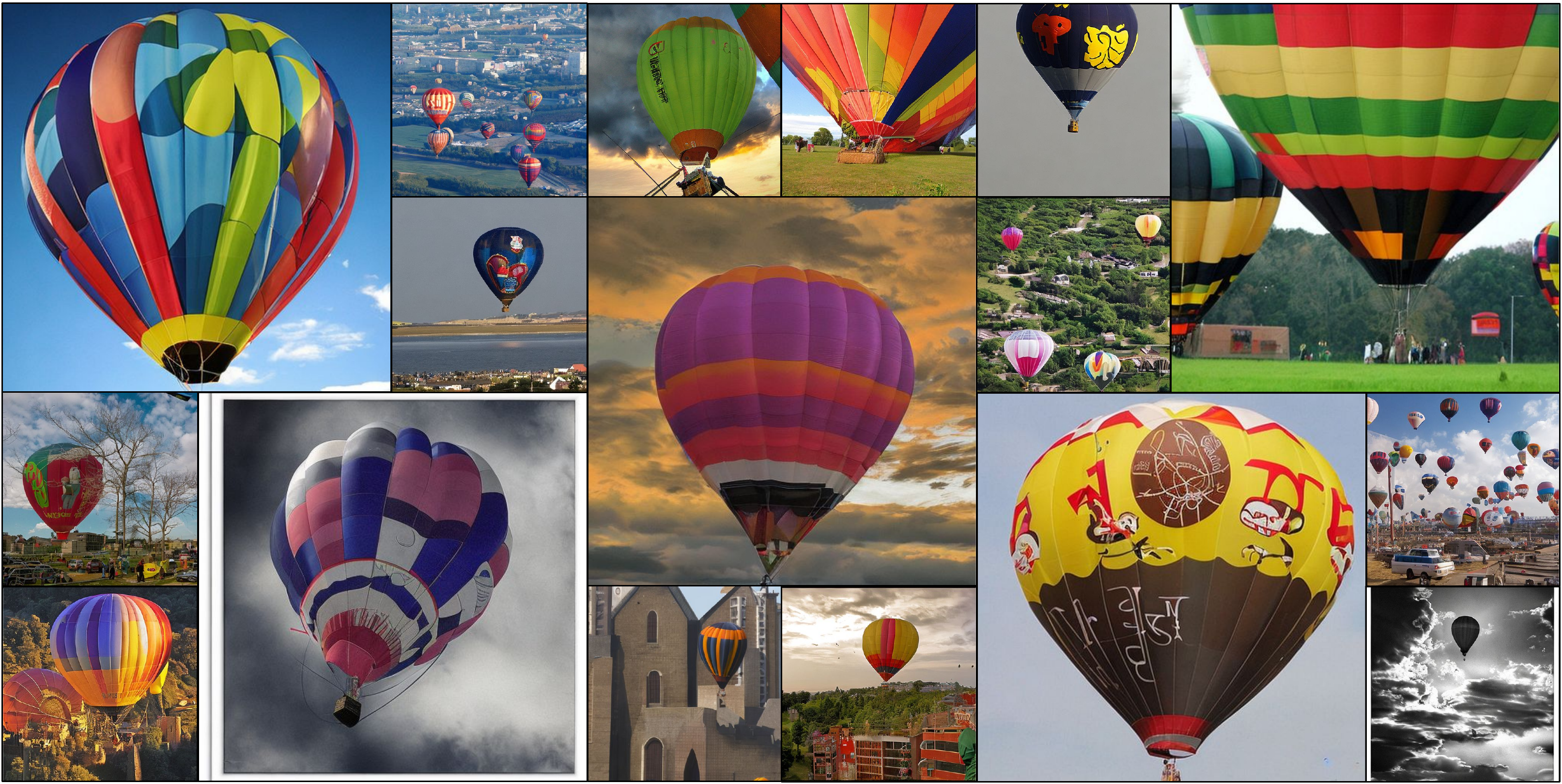}
    \caption{
    \textbf{Uncurated $512 \times 512$ \modelname samples.} Class label = `balloon' (417)}
    \vspace{-3ex}
    \label{fig:supp_512_6}
\end{figure*}

\begin{figure*}
    \centering
    \includegraphics[width=\linewidth]{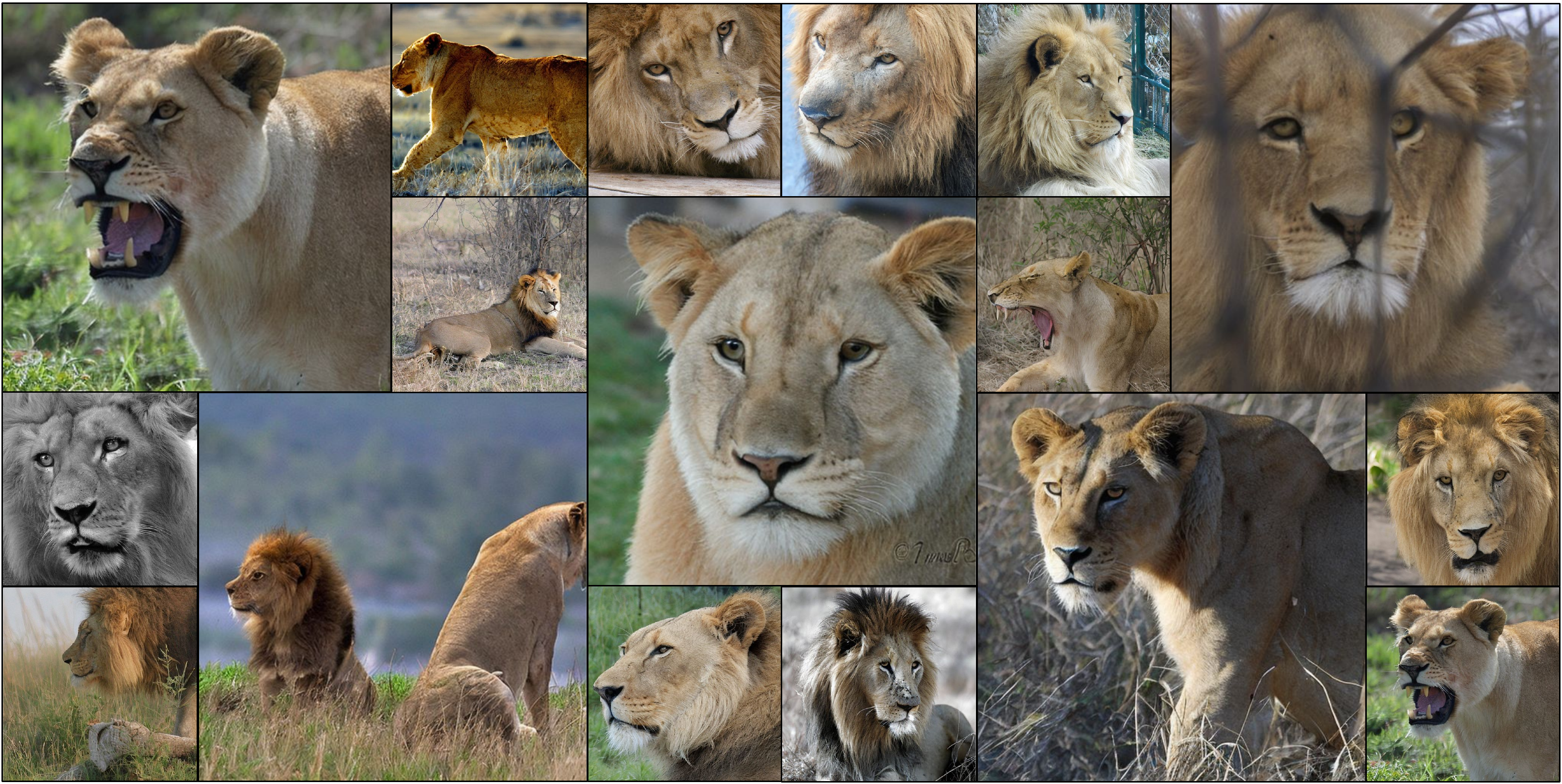}
    \caption{
    \textbf{Uncurated $512 \times 512$ \modelname samples.} Class label = `lion' (291)}
    \vspace{-3ex}
    \label{fig:supp_512_7}
\end{figure*}

\begin{figure*}
    \centering
    \includegraphics[width=\linewidth]{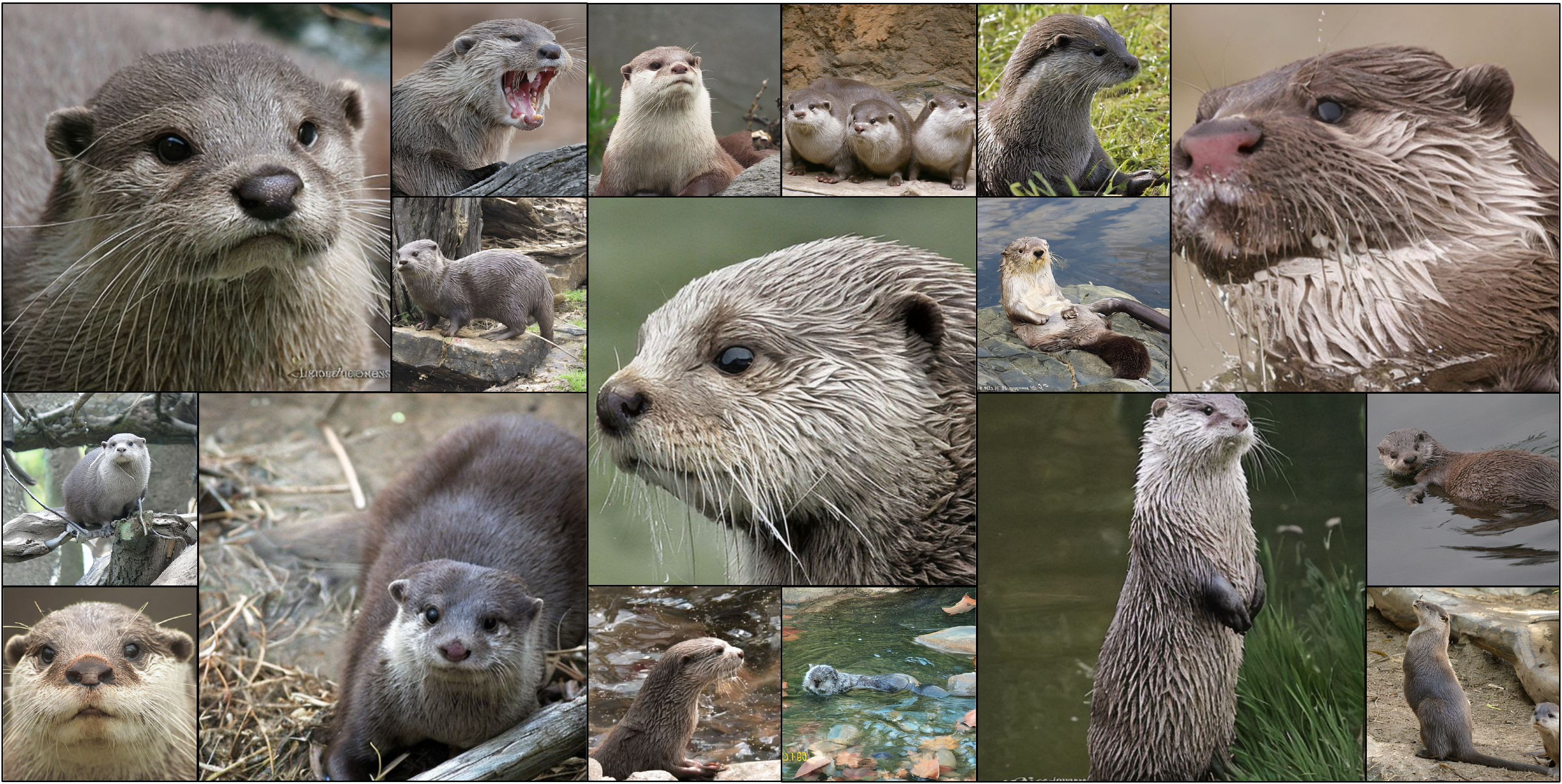}
    \caption{
    \textbf{Uncurated $512 \times 512$ \modelname samples.} Class label = `otter' (360)}
    \vspace{-3ex}
    \label{fig:supp_512_8}
\end{figure*}

\begin{figure*}
    \centering
    \includegraphics[width=\linewidth]{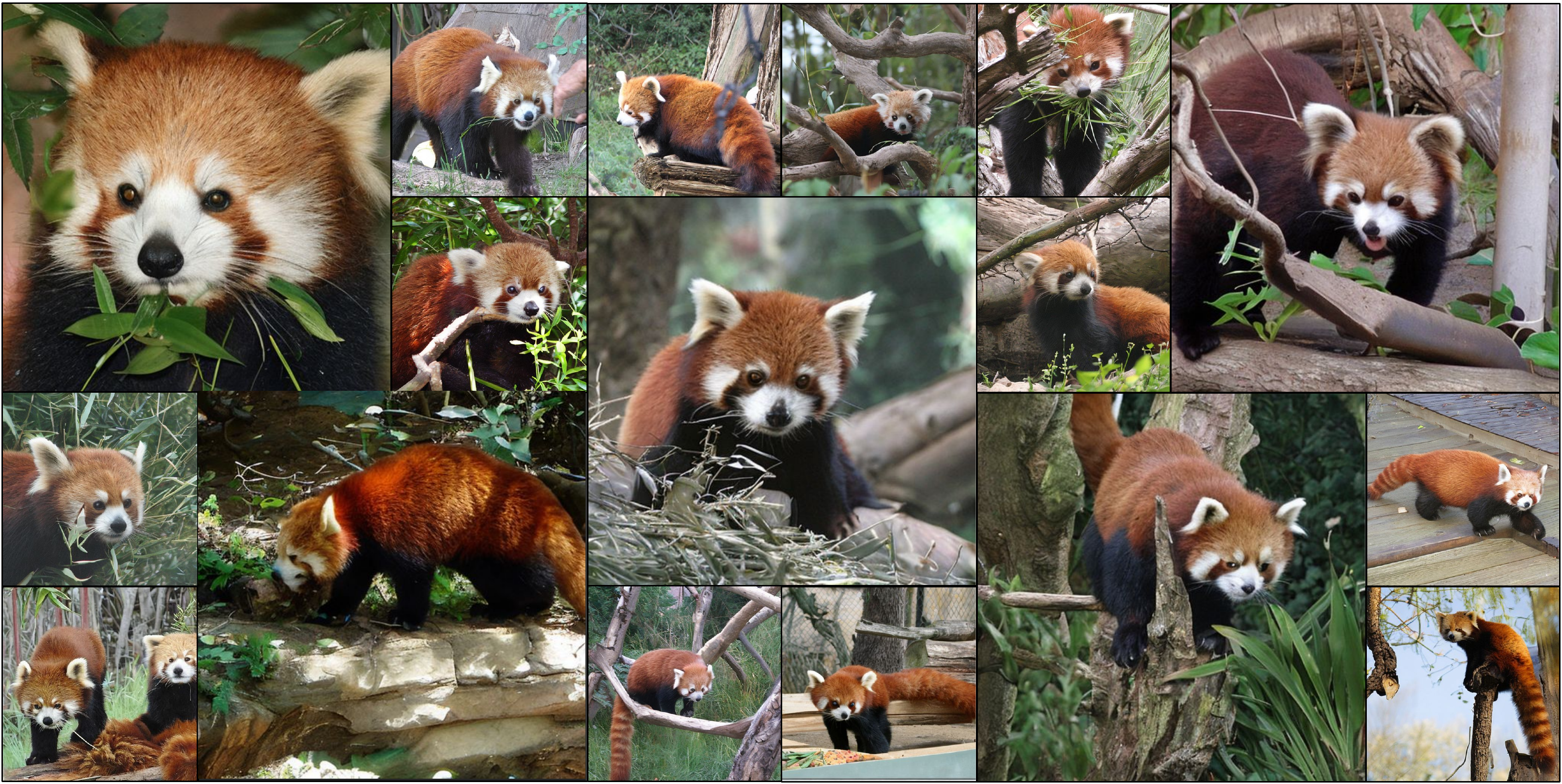}
    \caption{
    \textbf{Uncurated $512 \times 512$ \modelname samples.} Class label = `red panda' (387)}
    \vspace{-3ex}
    \label{fig:supp_512_9}
\end{figure*}

\begin{figure*}
    \centering
    \includegraphics[width=\linewidth]{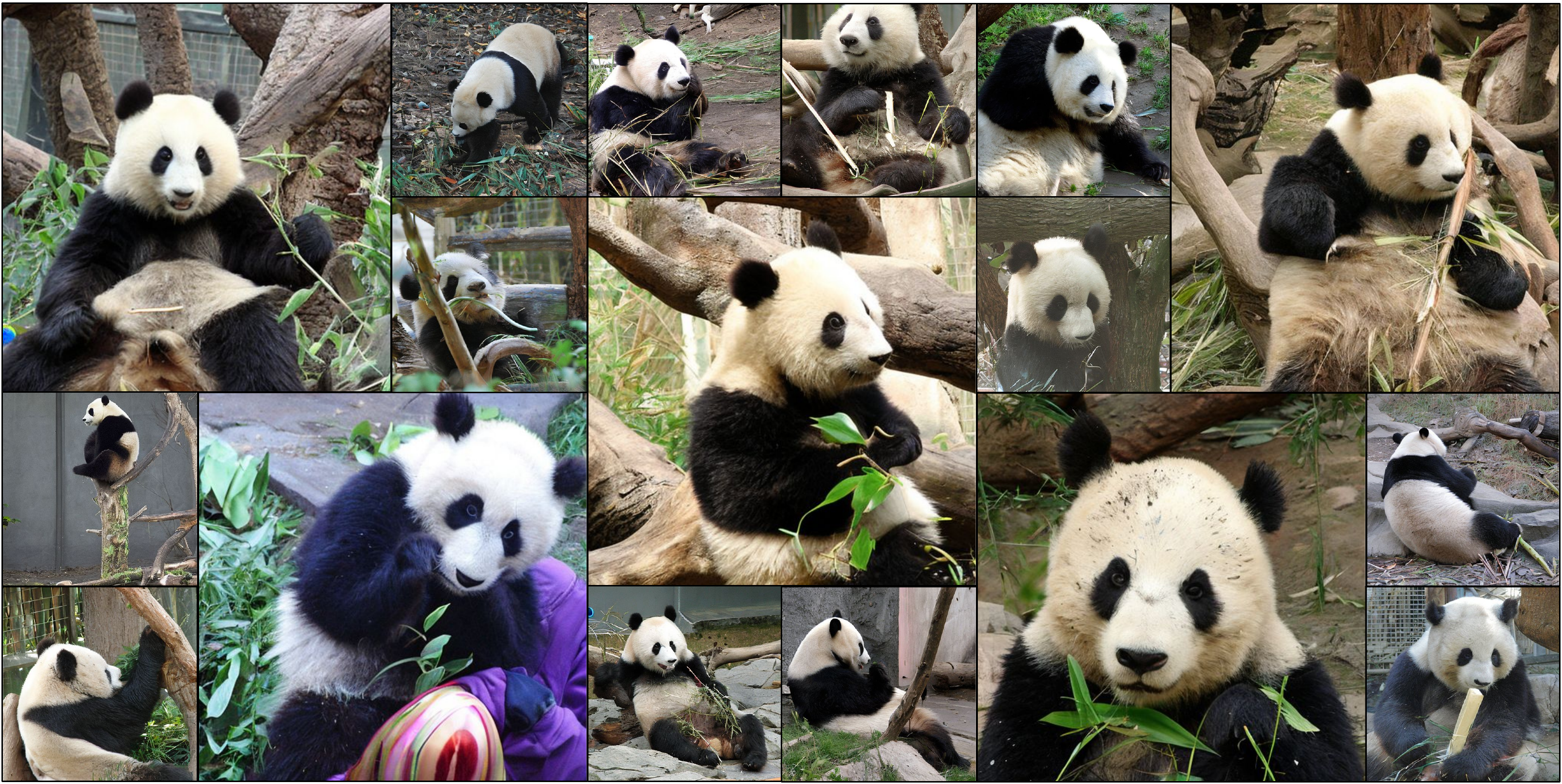}
    \caption{
    \textbf{Uncurated $512 \times 512$ \modelname samples.} Class label = `panda' (388)}
    \vspace{-3ex}
    \label{fig:supp_512_10}
\end{figure*}

\begin{figure*}
    \centering
    \includegraphics[width=\linewidth]{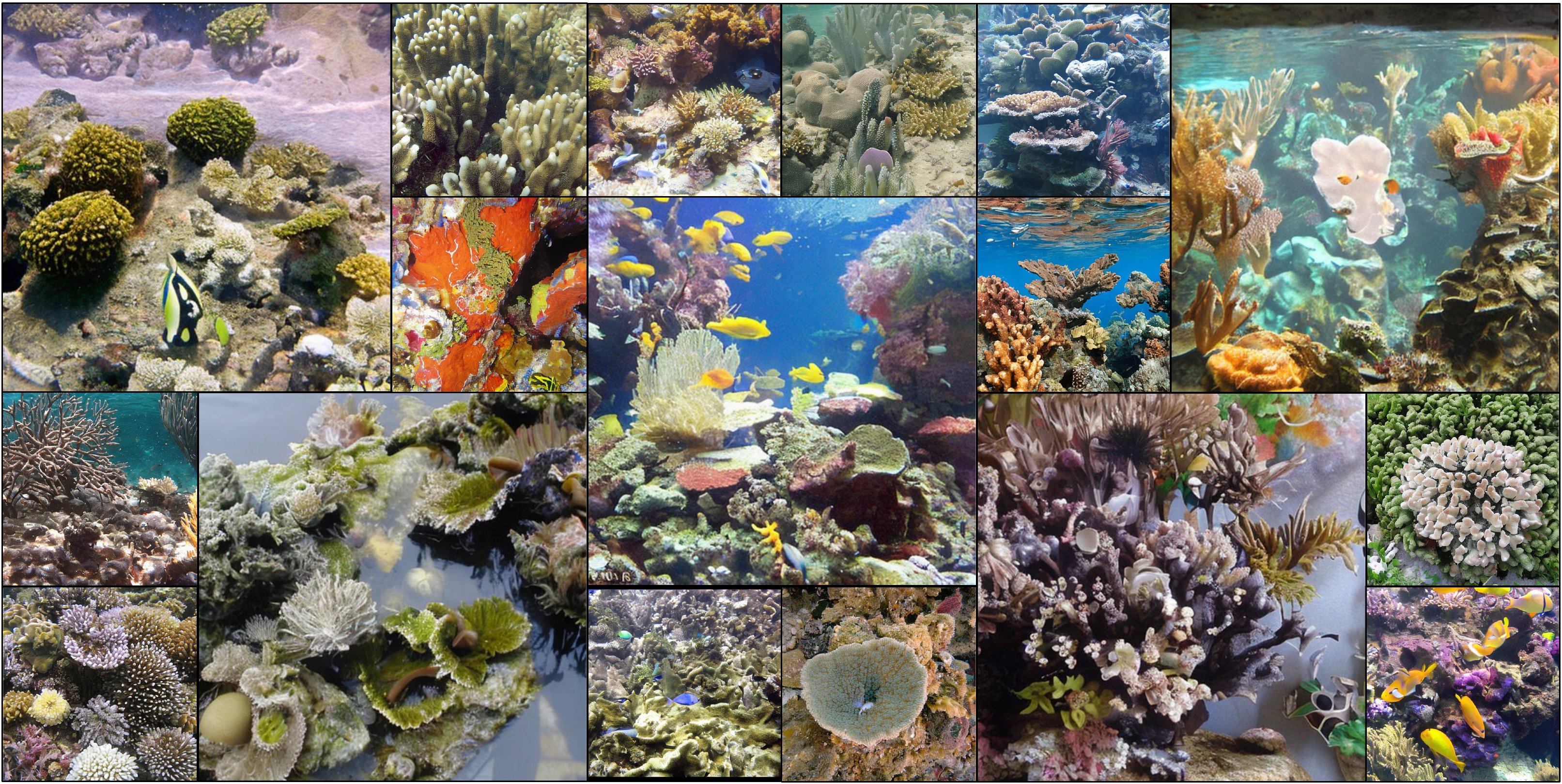}
    \caption{
    \textbf{Uncurated $512 \times 512$ \modelname samples.} Class label = `coral reef' (973)}
    \vspace{-3ex}
    \label{fig:supp_512_11}
\end{figure*}

\begin{figure*}
    \centering
    \includegraphics[width=\linewidth]{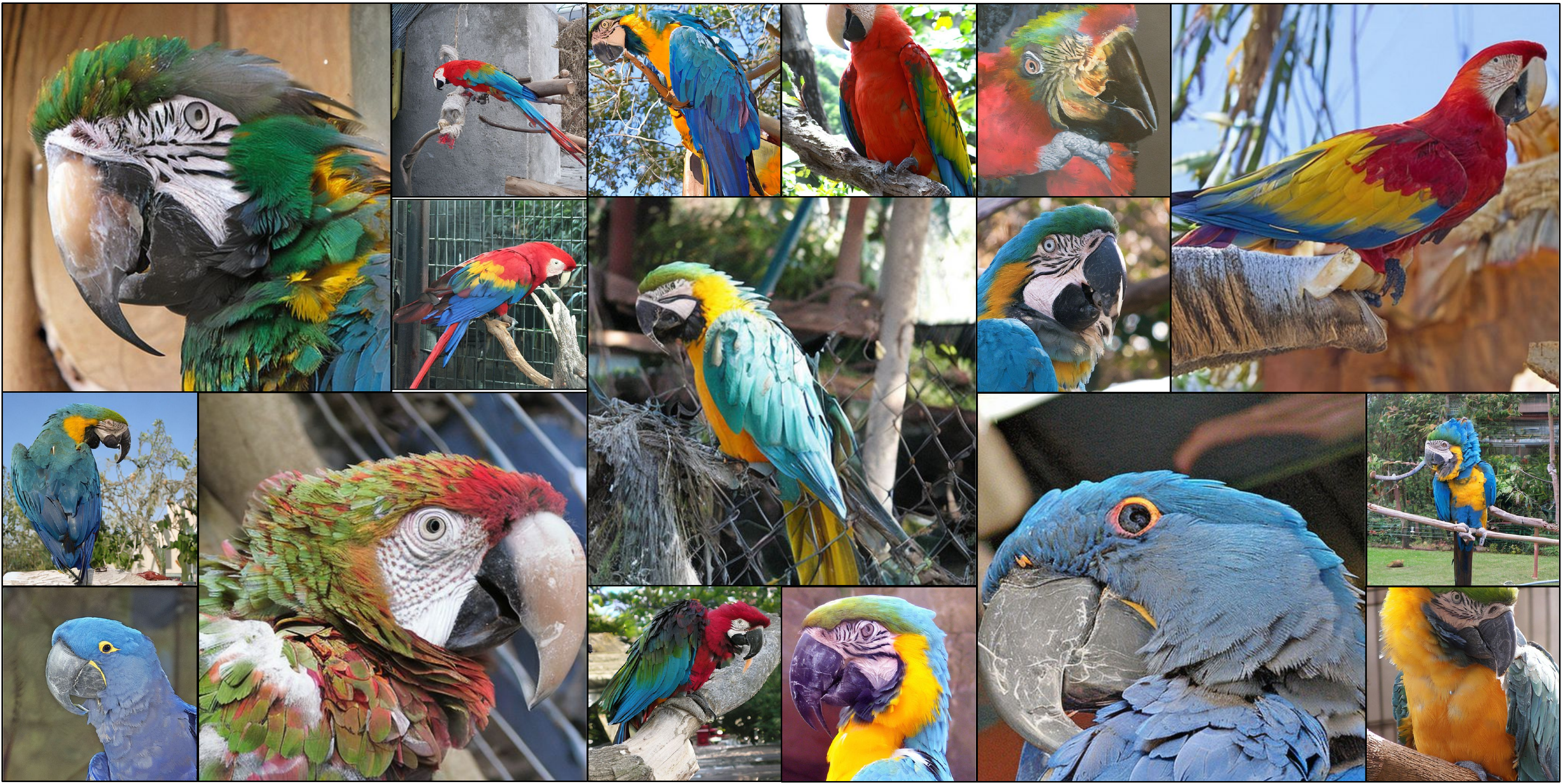}
    \caption{
    \textbf{Uncurated $512 \times 512$ \modelname samples.} Class label = `macaw' (88)}
    \vspace{-3ex}
    \label{fig:supp_512_12}
\end{figure*}

We present samples from our largest variant, \modelname-XL/3R, at 512 × 512 resolution trained for 800 epochs. Fig.~\ref{fig:supp_512_1}-~\ref{fig:supp_512_12} display uncurated samples from the model across a range of input class labels with classifier-free guidance. It is worth noting that our generated image samples exhibit high-quality and minimal image distortions.

\section{Limitations}
\label{sec:limitations}
The proposed \modelname has a few remaining limitations. First, it focuses on class-conditional image generation, rather than full text-to-image generation approaches. Additionally, although \modelname demonstrates great scalability in our experiments, the exploration of \modelname variants stops at the \modelname-XL/3R model with 524.8M parameters due to constraints on computational resources. In contrast, a few recent methods have scaled image diffusion models to billions of parameters. We leave it as future work to further explore the scaling law of our proposed \modelname to enhance its image generation capabilities.

\section{Broader Impacts}
\label{sec:impacts}
The proposed \modelname has the potential to facilitate numerous fields through its advanced image generation capabilities. In the realm of creative industries, \modelname can enhance the efficiency and creativity of artists and designers by generating high-fidelity images with fewer distortions. The high-quality generated images can also contribute to research on synthetic datasets by creating realistic images, aiding in reducing the annotations required for training vision models. However, with these advancements come ethical considerations, such as the risk of generating deepfakes or other malicious content. It is thus crucial to implement safeguards to minimize potential harms.

\section{Safety Concerns and Safeguards}
\label{sec:safe}
Given the powerful capabilities of \modelname, it is essential to implement robust safeguards to address potential safety and ethical concerns. One primary concern is the misuse of generated content, such as the creation of deepfakes, which can lead to misinformation and privacy violations. To mitigate this, it is important to establish strict access controls and usage policies to prevent the misuse of these models when released. Transparency in the training data and model architecture is also critical to ensure accountability and to identify potential biases that could lead to harmful outputs. By prioritizing these safeguards, we can ensure the responsible use of \modelname while minimizing potential risks.




\end{document}